\documentclass[review,authoryear]{elsarticle}

\usepackage{framed,multirow}

\usepackage{amssymb}
\usepackage{latexsym}
\usepackage{amsmath}
\usepackage{mathtools}
\usepackage{algorithm}
\usepackage{algpseudocode}
\usepackage{footnote}
\usepackage{placeins}
\usepackage{amsmath}
\usepackage{textcomp}
\usepackage{multirow}
\usepackage{placeins}
\usepackage{footnote}
\usepackage{tabulary}
\usepackage{graphicx}

\usepackage{subcaption}
\usepackage{caption}
\usepackage[table]{xcolor}
\usepackage{comment}
\usepackage{booktabs}
\usepackage{threeparttable}

\usepackage{hyperref}

\usepackage{makeidx,color,graphicx,algorithmicx,algorithm,algpseudocode,float}
\usepackage[table,xcdraw]{xcolor}
\usepackage{stfloats}
\usepackage{url}
\usepackage{xcolor}
\definecolor{newcolor}{rgb}{.8,.349,.1}

\usepackage{soul}

\usepackage[T1]{fontenc}

\usepackage[normalem]{ulem}
\definecolor{darkgreen}{rgb}{0,.4,0}
\definecolor{darkcyan}{rgb}{0,.4,.4}
\newcommand{\REMOVE}[1]%
          {{\color{blue}\sout{#1}}}

\newcommand{\COMMENT}[1]%
          {{\color{darkgreen}\textbf{{Editor: }} {#1}}}
\begin{document}
\journal{Remote Sensing of Environment}

\begin{comment}
\thispagestyle{empty}
\begin{table*}[!t]
\section*{Research Highlights}
\fboxsep=6pt
\fbox{
\begin{minipage}{.95\textwidth}
The main contributions of the article titled "Label-Efficient 3D Forest Mapping: Self-Supervised and Transfer Learning for Instance Segmentation, Semantic Segmentation, and Species Classification" are listed below
\vskip1pc
\begin{itemize}
 \item Unified 3D framework for forest instance, semantic, and species classification. 
 \item High performance using limited annotations and few-shot learning for unseen species.
 \item Systematic evaluation of self-supervised and transfer learning to reduce labels.
 \item Validated across diverse forest environments and realistic training scenarios.

\end{itemize}
\vskip1pc
\end{minipage}
}

\end{table*}
\end{comment}

% \clearpage

\pagenumbering{arabic}
\setcounter{page}{1}

\begin{frontmatter}

\title{Label-Efficient 3D Forest Mapping: Self-Supervised and Transfer Learning for Instance Segmentation, Semantic Segmentation, and Species Classification}

\author[1,2]{Aldino Rizaldy\corref{cor1}}
\ead{a.rizaldy@hzdr.de}
\author[3]{Fabian Ewald Fassnacht}
\author[1]{Ahmed Jamal Afifi}
\author[2]{Hua Jiang}
\author[1]{Richard Gloaguen}
\author[1,4]{Pedram Ghamisi}

\cortext[cor1]{Corresponding author: Aldino Rizaldy} 

\address[1]{Helmholtz-Zentrum Dresden-Rossendorf (HZDR), Helmholtz Institute Freiberg for Resource Technology (HIF), 09599 Freiberg, Germany.}
\address[2]{Remote Sensing and Geoinformatics, Freie Universität Berlin, 12249 Berlin, Germany}
\address[3]{Institute of Geomatics, BOKU University, 1190 Vienna, Austria}
\address[4]{Faculty of Electrical and Computer Engineering, University of Iceland, 101 Reykjavik, Iceland}

%\received{X XXX 20XX}
%\finalform{X XXX 20XX}
%\accepted{X XXX 20XX}
%\availableonline{X X 20XX}
%\communicated{XXX}

\begin{abstract}
Detailed structural and species information on individual tree level is increasingly important to support precision forestry, biodiversity conservation, and provide reference data for biomass and carbon mapping. Point clouds from airborne and ground-based laser scanning are currently the most suitable data source to rapidly derive such information at scale. Recent advancements in deep learning improved segmenting and classifying individual trees and identifying semantic tree components. However, deep learning models typically require large amounts of annotated training data which limits further improvement. Producing dense, high-quality annotations for 3D point clouds, especially in complex forests, is labor-intensive and challenging to scale. We explore strategies to reduce dependence on large annotated datasets using self-supervised and transfer learning architectures. Our objective is to improve performance across three tasks: instance segmentation, semantic segmentation, and tree classification using realistic and operational training sets. We observe improvements across all tasks, compared to training from scratch, evaluated with their respective metrics. For instance segmentation, self-supervised learning combined with domain adaptation improves AP50 by 16.98\%. For semantic segmentation, self-supervised learning alone improves mIoU by 1.79\%. For tree classification, hierarchical transfer learning improves mean Jaccard by 6.07\%. To simplify use and encourage uptake, we integrated the tasks into a unified framework, streamlining the process from raw point clouds to tree delineation, structural analysis, and species classification. Pretrained models reduce energy consumption and carbon emissions by ~21\%. This open-source contribution aims to accelerate operational extraction of individual tree information from laser scanning point clouds to support forestry, biodiversity, and carbon mapping. 
\end{abstract}
%The code is available on \url{https://github.com/aldinorizaldy/TreeLite3D}.

\begin{keyword}
Forest \sep self-supervised learning \sep transfer learning \sep domain adaptation \sep segmentation \sep classification \sep deep learning

%% MSC codes here, in the form: \MSC code \sep code
%% or \MSC[2008] code \sep code (2000 is the default)
\end{keyword}

\end{frontmatter}

%\linenumbers

\section{Introduction} 
\label{sec:intro}
Tree-level analysis of forest structures is critical for both ecological analyses of forests and to inform operational forest management (\cite{10.1093/forestry/cpad024, NAESSET200288}). Structural attributes such as diameter at breast height (DBH), tree height, and crown dimensions are essential for estimating key forest variables, such as basal area, biomass, and carbon (\cite{chave2014improved, salas2010modelling}). The measurements also support forest and ecosystem modeling and can serve as reference data to train regional or even global empirical models (\cite{cao2016aboveground}).

Individual-tree forest attributes are typically measured in field plots as part of national or local inventories. These surveys are limited in scope due to cost and time, often covering only a small number of plots (\cite{mcroberts2007remote}). While traditional field inventories have been the standard for decades, emerging remote and proximal sensing technologies, along with increasing demands for higher temporal resolution, are challenging this approach (\cite{finger2025comparison,tompalski2019challenges}).

A clear example of these challenges arises in the context of climate change, where increasing frequency and intensity of forest disturbances shorten the period for which inventory data remains up-to-date. Remote sensing offers a solution to provide more frequent updates, but high-resolution models still depend on calibration data from field surveys, which may be unavailable after disturbances (\cite{senf2017remote}). With improving Airborne Laser Scanning (ALS) quality, the need for calibration data from the field may be strongly reduced or even become obsolete in the future. Individual-tree information extracted from ALS data, in combination with simple allometries, can produce high-quality estimates of various forest inventory parameters (\cite{LATIFI2015162,hyyppa2012advances}). However, one remaining challenge for such approaches is the correct identification and delineation of individual trees to which the allometries are then applied.

Besides the timeliness of reference data, data quality is increasingly seen as an important issue. The limited quality of field measurements has been particularly discussed in the context of remote-sensing-assisted global biomass mapping and estimation efforts (\cite{vorster2020variability}). Any remote sensing model cannot be better than the quality of the reference data it is calibrated with. For example, biomass reference data derived using allometries such as traditional field measurements (height, DBH) and species can entail important sources of uncertainty. Individual tree biomass estimates can easily vary between 10-20 percent depending on the allometry applied (\cite{duncanson2017implications}). Here, data from Terrestrial Laser Scanning (TLS), Mobile Laser Scanning (MLS), and Drone Laser Scanning (DLS) systems could potentially lead to large improvements by enabling a precise estimate of the tree volume which can then be translated into a woody biomass or carbon estimate either directly (using a species-specific wood density) or using simple empirical models between the tree volume and biomass with the biomass values obtained from a limited number of destructive samples. In both cases, the estimation of the individual tree volume requires the accurate extraction of the woody compartments of the tree from point clouds. While various approaches, for example, Quantitative Structural Models (QSM), have been introduced over the last years (\cite{brede2019non}), the task remains challenging, particularly in complex forest stands (\cite{morhart2024limitations}).

Recent advances in deep learning and computer vision methods led to the development of powerful methods for extracting detailed structural and semantic information from complex 3D forest environments (\cite{HENRICH2024102888, XIANG2024114078, WIELGOSZ2024114367}). Despite notable progresses, a fundamental challenge remains: \textbf{deep learning models require a large volume of annotated data to perform effectively} (\cite{krizhevsky2012imagenet}). 

In dense forest environments, annotating high-density 3D point clouds is particularly difficult due to occlusions, overlapping tree structures, and variable tree types (\cite{Puliti2023FORinstance}). Manual annotation is time-consuming and labor-intensive, thus impractical for scaling up to larger or diverse datasets, especially at the instance level, as the distinction of individual trees is highly complex. 

While open datasets exist (\cite{Puliti2023FORinstance, Puliti2025}), their sizes and diversities are often limited. This makes them less suitable for practitioners applying deep learning methods to new forest areas. There is an urgent need for learning strategies that can perform effectively under limited supervision and where the creation of dense labels is challenging. Few-shot learning approaches have emerged as a promising solution, enabling the models to learn from a small number of labeled samples while maintaining strong performance across segmentation and classification tasks.

Self-supervised learning (SSL), particularly contrastive learning, offers a way to leverage large volumes of unlabeled point cloud data (\cite{xie2020pointcontrast,hou2021contrastive,wu2023masked}). SSL extracts general-purpose features by exploiting structural similarities within the data without requiring manual annotation. In this study, we propose to pretrain a deep network using contrastive SSL and then fine-tune it for instance and semantic segmentation tasks bydemonstrating that SSL features are transferable and effective under limited supervision.

However, SSL-derived features may not always capture task-specific information on the target domain (\cite{saito2020universal}). This is particularly true for instance segmentation, where we observe that the task requires learning instance-level distinctions. To address this, we incorporate domain adaptation techniques on top of the pretrained SSL backbone. By aligning feature distributions between source and sparsely labeled target data, domain adaptation enhances instance segmentation accuracy in new forest areas.

For classification, we suggest to employ hierarchical transfer learning. Models are first pretrained to classify trees into broad categories (e.g., coniferous vs. broadleaf) to capture coarse semantic information and then fine-tuned for species-level classification. This hierarchical approach leverages the structure of botanical taxonomy to improve fine-grained performance with limited labeled data.

These strategies complement each other by addressing data scarcity at different levels. SSL provides a label-free geometric foundation for point-wise features, while domain adaptation compensates for the lack of instance-level grouping features in contrastive SSL, by aligning clustering logic across domains. This extraction of individual trees allows instance-level species classification in forest plot data, where hierarchical transfer learning leverages taxonomic structures to identify specific species from limited samples. This integrated flow ensures that the model moves from understanding raw geometric points to species identity.

In this study, we demonstrate that our approach achieves strong performance in few-shot learning settings. Notably, for instance segmentation, our method can segment individual trees using only 0.01\% labeled points (approximately 4-5 points per tree), whereas training from scratch fails. By integrating SSL, domain adaptation, and hierarchical transfer learning into a unified framework, we provide a scalable workflow for converting raw point clouds into individual tree delineation, structural analysis, and species classification, even in label-constrained environments. 

The key contributions of our study are:
\begin{itemize}
    \item A unified 3D deep learning framework that integrates state-of-the-art pretraining, domain adaptation, and hierarchical transfer learning for forest analysis.
    \item Pretraining on a large, heterogeneous collection of point clouds covering diverse forest types, species, and acquisition modalities, ensuring broad generalization.
    \item Strong performance under limited training data, addressing few-shot learning challenges in forestry.
    \item Extensive validation across multiple sites, demonstrating practical relevance for stakeholders deploying deep learning in forest monitoring.
\end{itemize}

%%%%%%%%%%%%%%%%%%%%%%%%%%%%%%%%%%%%%
\section{Related work}
\label{sec:related}

\subsection{Existing tree segmentation and classification methods}

Tree segmentation and classification can be approached from either \textit{image-based} or \textit{point-based} methods. Image-based approaches operate on RGB or multispectral imagery, while point-based methods leverage 3D data from airborne, drone, or terrestrial lidar. Point-based methods generally offer higher geometric accuracy, reduced occlusion ambiguity, and the ability to capture tree structure directly on point-level without requiring a typical projection of points to Canopy Height Models (CHMs), making them advantageous in dense or heterogeneous forests.

Traditional image-based segmentation often employs handcrafted features and spatial analysis, such as minimum spanning tree segmentation, watershed transforms, and object-based image analysis (\cite{BLASCHKE2014180, QIN2022113143}). Deep learning methods have improved robustness in canopy delineation and species classification by leveraging object detectors and semantic segmentation networks (\cite{Weinstein2020DeepForest, Ball2023Detectree2, Chen2025ZeroShotTree}).

 There has been a surge in methods to extract detailed tree-level insights from 3D point clouds in forest environments in recent years. Early approaches relied on geometric analysis of canopy height models, marker-controlled watershed segmentation, and clustering techniques to delineate individual trees from lidar data (\cite{Li2012Segmenting, DalponteCoomes2016, Chen2006Isolating, Silva02092016, ROUSSEL2020112061, 9033973, drones8120772}). While computationally efficient, these methods often fail in complex canopy situations due to occlusion and structural complexity in dense or overlapping canopies (\cite{HENRICH2024102888}).

To address these challenges, researchers have developed deep learning techniques. Architectures such as PointGroup (\cite{Jiang2020PointGroup}), which have later been adopted in recent studies (\cite{XIANG2024114078, WIELGOSZ2024114367, HENRICH2024102888}), are now enabling robust instance segmentation by learning context-aware features from raw point clouds. Similarly, semantic segmentation has advanced from geometric and handcrafted feature classifiers to deep point- and voxel-based approaches like Pointnet++ and sparse 3D Unet to improve classification of branches, foliage, and stems (\cite{rs15194793, XIANG2024114078}). Tree species and structural classification have also benefited from both image- and point-based approaches (\cite{HAMRAZ2019219, BOLYN2022113205, rs17071190}). 

Deep learning has also been applied for tree species classification (\cite{liu2022tree,vahrenhold2025mmtscnet,fan2023tree}). The recent FOR-species20K dataset (\cite{Puliti2025}) further demonstrates how large-scale data and deep models enable species recognition across diverse tree types. Through this progression, forestry applications evolve from heuristic and labor-intensive workflows to scalable, data-driven methods. However, most deep learning approaches still rely on densely labeled data, and techniques for few-shot learning scenarios remain largely unexplored.

\subsection{Learning with limited supervision}
Parallel to forestry advances, computer vision has significantly influenced how models can be trained with limited or no labels. SSL has emerged as a powerful tool to learn transferable representations from unlabeled data. Early pioneering models include Simple Framework for Contrastive Learning of Visual Representations (SimCLR) (\cite{chen2020simple}), which uses contrastive learning with strong data augmentations to bring similar images closer in feature space; Momentum Contrast (MoCo) (\cite{he2020momentum}), which introduces a momentum-based dynamic dictionary to efficiently scale contrastive learning; and Masked AutoEncoder (MAE) (\cite{he2022masked}), which randomly masks a large portion of image patches and trains a lightweight decoder to reconstruct the missing content, enabling the model to learn rich contextual representations.

In the 3D domain, similar strategies have been adapted to exploit the structure and geometry of point clouds. PointContrast (\cite{xie2020pointcontrast}) applies contrastive learning on paired 3D patches from different views of the same scene, encouraging viewpoint- and density-invariant representations. Contrastive Scene Contexts (CSC) (\cite{hou2021contrastive}) captures spatial and contextual relationships between local 3D patches, using contrastive objectives to maintain neighborhood consistency. Masked Scene Contrast (MSC) (\cite{wu2023masked}) extends masked modeling to 3D, masking portions of a point cloud and reconstructing them while enforcing contrastive consistency across scenes. Finally, Point-BERT (\cite{yu2022pointbert}) adapts the masked language modeling paradigm to 3D, tokenizing point clouds into discrete “point tokens” and predicting masked tokens, which strengthens both local and global geometric understanding. These approaches allow 3D models to generalize effectively across scenes, tasks, and datasets, even with sparse labeled data.

Transfer learning further allows models pretrained on large datasets to be adapted to new tasks with limited labeled data. Deep Convolutional Activation Feature (DeCAF) (\cite{donahue2014decaf}) showed that features from deep Convolutional Neural Networks (CNNs) trained on ImageNet provide effective generic representations, while \cite{yosinski2014transferable} highlighted that lower-layer features are more general and higher-layer features are more task-specific, guiding effective fine-tuning strategies.

Domain adaptation addresses shifts between source and target distributions. Domain Adversarial Training of Neural Networks (DANN) (\cite{ganin2016domain}) uses adversarial training to align feature distributions, and Adversarial Discriminative Domain Adaptation (ADDA) (\cite{tzeng2017adversarial}) adapts target features with a separate encoder and discriminator. These methods enable models to generalize across different datasets, which is crucial for applications like forestry, where environments and sensors vary.

Despite progress in both domains, workflows developed in a forestry context so far rarely capitalize on structured SSL, transfer learning, and domain adaptation, and computer vision methods seldom target forestry-specific challenges. Our work aims to bridge this gap by integrating contrastive pretraining, domain adaptation, and hierarchical transfer into a unified 3D learning framework for simultaneous instance segmentation, semantic segmentation, and tree classification, empowering models to perform effectively even when labeled data is scarce.
%%%%%%%%%%%%%%%%%%%%%%%%%%%%%%%%%%%%%
\section{Dataset and Methodology}
\label{sec:methodology}

\subsection{Dataset}
\label{sec:dataset}

\textbf{Benchmark datasets.} We evaluate our framework using two benchmark datasets: FOR-instance (\cite{Puliti2023FORinstance}) and FOR-species20K (\cite{Puliti2025}). The FOR-instance dataset is a machine learning-ready benchmark specifically designed for advancing instance and semantic segmentation techniques of individual trees from dense drone-borne laser scanning data. The dataset comprises five curated lidar data collections from diverse locations, including Norway (NIBIO), the Czech Republic (CULS), Austria (TU-WIEN), New Zealand (SCION), and Australia (RMIT). The dataset features manual annotation for individual trees (instance) and semantic classes, such as stem, woody branches, live branches, low vegetation, and terrain. It represents complex forest types, ensuring broad capability and transferability of developed models.

%We follow the official plot-level partitioning scheme provided by the FOR-instance benchmark. The data is divided into a development set and a test set. Following \cite{XIANG2024114078}, we further subdivided the development set into 42 training plots and 14 validation plots. The 11 test plots remained held for final evaluation. 

The FOR-species20K dataset is a large, publicly available benchmark for individual tree species classification using laser scanning data. It consists of 20,158 individual tree point clouds representing 33 distinct tree species. More than 70\% of the point clouds were collected by terrestrial laser scanning, with the remainder sourced from drone-borne and mobile laser scanning. The data primarily comes from European forest ecosystems, covering temperate, boreal, and Mediterranean regions.

\textbf{Unlabeled data for pretraining.} A diverse collection of unlabeled 3D point cloud data was compiled from publicly available sources \footnote{U.S. Geological Survey, ``3D Elevation Program (3DEP),'' 2023, \url{https://www.usgs.gov/3dep}} \footnote{Het Waterschapshuis, ``Actueel Hoogtebestand Nederland (AHN-4),'' 2022, \url{https://www.ahn.nl/}} \footnote{Estonian Land and Spatial Development Board, ``Airborne LiDAR (ALS) Point Cloud Data of Estonia,'' 2024, \url{https://geoportaal.maaamet.ee/}} \footnote{British Columbia Ministry of Water, Land and Resource Stewardship, ``LidarBC,'' 2024, \url{https://lidar.gov.bc.ca}} \footnote{Government of New Brunswick, ``Lidar Data,'' 2023, \url{https://www.gnb.ca/en/campaign/geonb/data-catalogue/lidar-data.html}} and collaborating research institutions, encompassing multiple countries, forest types, geographic conditions, and acquisition modalities. The main characteristics of the unlabeled data are summarized in Table \ref{tab:unlabeled_data}.

\begin{table}[htbp]
\centering
\caption{Summary of unlabeled LiDAR datasets used for contrastive pretraining.}
\label{tab:unlabeled_data}
\begin{tabular}{p{0.98\linewidth}}
\toprule
\textbf{Droneborne LiDAR} \\
\hspace{1em}Number of tiles: $\sim$400 \\
\hspace{1em}Total points: $\sim$800 million \\
\hspace{1em}Tile size: 20 $\times$ 20 m \\
\hspace{1em}Average point density: $\sim$5,000 pts/m$^2$ \\
\hspace{1em}Total area: $\sim$16 ha \\
\hspace{1em}Region: Germany \\
\hspace{1em}Environment: Dense canopy, fine-scale structural detail \\
\textbf{Airborne LiDAR} \\
\hspace{1em}Number of tiles: $\sim$5,500 \\
\hspace{1em}Total points: $\sim$1.5 billion \\
\hspace{1em}Tile size: 100 $\times$ 100 m \\
\hspace{1em}Average point density: 20-30 pts/m$^2$ \\
\hspace{1em}Total area: $\sim$5,500 ha \\
\hspace{1em}Region: Multi-country (Europe, North America, South America) \\
\hspace{1em}Environment: Mixed forest types (temperate, boreal, Mediterranean, and tropical), varied topography \\
\bottomrule
\end{tabular}
\end{table}

% Given the significant variation of point density between different acquisition system, specifically Airborne Laser Scanning (ALS) typically produces much lower point density compared to Drone Laser Scanning (DLS), MLS, or TLS, we pretrain two separate backbone models, one model specialized for ALS data and another model for denser point clouds from DLS, MLS, and TLS sources.

% This separation ensures that the backbone is adapted to the structural characteristics and resolution to each sensing modality.

%\subsection{Data preprocessing}

Standard preprocessing involved recentering XY coordinates and normalizing intensity and echo values to the [0,1] range. To accommodate large scenes, the data were tiled into 20 x 20 m patches for droneborne LiDAR and 100 x 100 m for Airborne LiDAR.

\subsection{Overview of the framework}
This work investigates the effectiveness of combining self-supervised pretraining with supervised adaptation for multi-level forest scene understanding using 3D point clouds. Our study focuses on three main tasks: (1) instance segmentation, (2) semantic segmentation, and (3) individual tree classification. Following the idea of foundation models in earth observation, such as SpectralGPT (\cite{10490262}), the key idea is to first learn general-purpose point-level features through contrastive self-supervised learning. These pretrained representations are then leveraged in two ways: either by directly fine-tuning the encoder for downstream tasks, or by integrating task-specific heads and performing supervised training on the labeled source domain before fine-tuning the entire model on the target domain. 

To enable generalization across diverse forest types with limited annotation, we incorporate a domain adaptation strategy into our training pipeline. Specifically, we first adapt the pretrained encoder on a fully labeled source region, then fine-tune it on a new target region with only sparse supervision. This setting reflects a realistic application scenario where labeled data is scarce in operational conditions. 

In both scenarios, we evaluate the framework under varying degrees of labeled supervision to assess its ability to perform in a few-shot learning setting. Figure \ref{fig:overview} illustrates the framework along with the three downstream tasks including:

\begin{figure*}[ht!]
    \centering
    \includegraphics[width=0.98\linewidth]{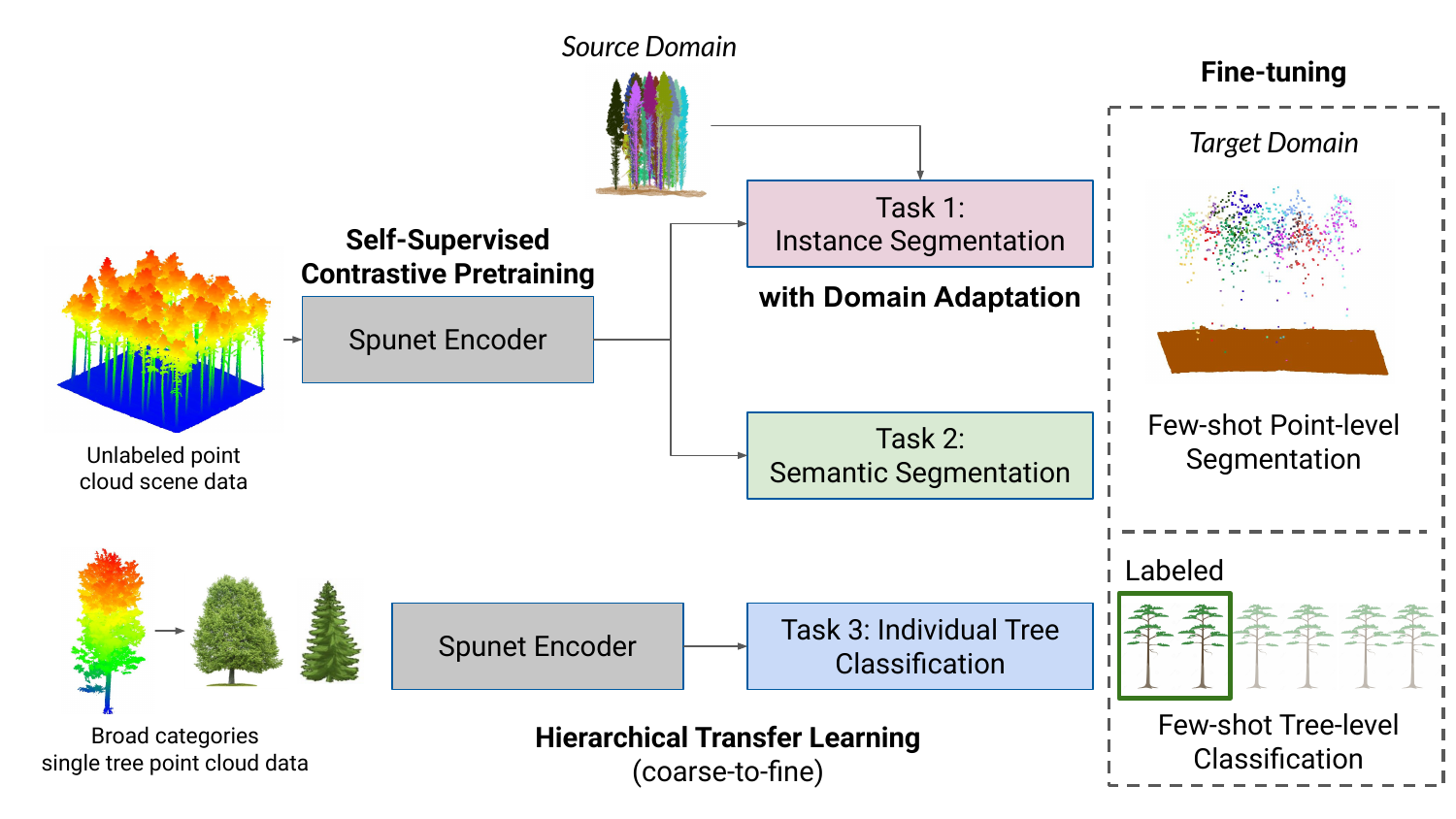}
    \caption{Overview of the framework. }
    \label{fig:overview}
\end{figure*}

%Our framework consists of two main parts: the pretrained backbone network and the heads. The former serves as a general features extractor, while the latter serves in a specific task. We train our framework in two stages: the first stage is unsupervised training using SSL and the second stage is supervised training, both with full supervision and data-efficient learning to observe the performance of the model in different scenarios. Fig. illustrates the overall framework and the strategies we use in this study. 

An \textbf{Instance segmentation branch} which aims to detect and separate individual tree instances from a 3D forest point clouds by predicting offsets and grouping points that belong to the same instance. 

A \textbf{Semantic segmentation branch} which assigns a class label to each point in the point cloud, enabling detailed per-point understanding of tree structure.

An \textbf{Individual tree classification branch }which predicts the species or types of each segmented tree instance by aggregating features and performing instance-level classification.

\subsection{Self-supervised pretraining}
We investigate the effectiveness of pretraining to learn rich, generalizable point-level features from unlabeled 3D point clouds of forest scenes, enabling effective transfer to downstream tasks. To do this, we utilize a self-supervised encoder pretrained using a contrastive learning strategy, specifically the Masked Scene Contrast (MSC) (\cite{wu2023masked}) pipeline. We adopt MSC because it operates directly on raw, scene-level 3D point clouds, unlike earlier methods that rely on frame RGB-D inputs (\cite{xie2020pointcontrast}). The design enables efficient and scalable pretraining across large point cloud datasets. 

Contrastive learning works by generating diverse views of the same data, encoding these views into feature representations in a backbone network, and matching points between views. Positive samples are identified based on the spatial proximity between different views of the matching points. The contrastive loss encourages similarity among positive pairs, while maintaining dissimilarity among negative pairs. This process leads the encoder to capture semantically and structurally meaningful features that do not rely on any manual annotations. The learned feature representations serve as the initialization for downstream tasks, where either direct fine-tuning or supervised adaptation can be applied depending on the use case. 

%We then investigate whether self-supervised pretraining alone is sufficient, by extending the pretrained weights on a supervised training on a source domain to learn more specific tasks, before deploying the model on the target domain, particularly for the limited supervision scenario. 

\textbf{Objective.} The goal of this pretraining phase is to learn a latent space by bringing representations of similar point neighborhoods (positive) closer in feature space while pushing dissimilar neighborhoods (negative) apart. We accomplish this by minimizing InfoNCE (Noise Contrastive Estimation) loss (Eq. \ref{eq:infonce}), which forces the model to effectively maximize the mutual information between the different augmented views of the same point cloud sample:

\begin{equation}
\mathcal{L}_{\text{InfoNCE}} = -\log \frac{\exp(\mathrm{sim}(z_i, z_j) / \tau)}{\sum_{k=1}^{N} \exp(\mathrm{sim}(z_i, z_k) / \tau)}
\label{eq:infonce}
\end{equation}

where $z_i$ and $z_j$ denote the embeddings of positive pairs, $\mathrm{sim}(\cdot)$ represents cosine similarity, $\tau$ is a temperature parameter controlling distribution sharpness, and the denominator sums over one positive and $N-1$ negative samples.

%By doing so, the model acquires a robust inductive bias about forest geometry, which is critical for achieving data efficiency in downstream learning. 

%\textbf{Augmentation strategy for contrastive learning}

\textbf{Backbone architecture.} We utilize a 3D network that follows the U-Net topology (\cite{cciccek20163d}) as the backbone encoder. This specific implementation, which we refer to as Sparse UNet, is adapted to efficiently handle large, sparse 3D volumetric data (such as point clouds) by computing only on active, non-empty voxels. This efficiency is achieved by leveraging sparse convolutions (\cite{graham20183d,choy20194d}), which makes the architecture scalable, fast, and robust for 3D data. Combined together, it enables efficient processing while capturing high-level features alongside fine-grained spatial details. 

The backbone Sparse UNet configuration is detailed as follows:

\begin{itemize}
    \item \underline{Voxelization}: All input point clouds are voxelized with a grid size of 5 cm to balance geometric detail and computational speed.
    \item \underline{Input specifications}: The network accepts 6-channel input features, i.e., three spatial coordinates (X,Y,Z), lidar intensity, return number, and number of returns.
    \item \underline{Encoder-decoder depth}: The network employs a 4-hierarchical structure. The encoder progressively downsamples the spatial resolution while increasing feature depth, followed by a symmetric decoder that restores spatial detail through skip connections.  
    \item \underline{Feature channel configuration}: The channel dimensions for the encoding stages are (32, 64, 128, 256), with the decoding stages following the sequence (256, 128, 96, 96).
    \item \underline{Block depth}: Each hierarchical level consists of multiple residual blocks. The number of layers per level is configured as (2,3,4,6) for the encoder and (2,2,2,2) for the decoder.
    \item \underline{Output dimensionality}: The final backbone output provides a 96-dimensional point-wise feature representation used for downstream task heads. 
\end{itemize}

%The architecture follows the encoder-decoder structure of the traditional U-Net [cite], using downsampling layers to capture high-level features and upsampling layers with skip connections to retain spatial details.

%\textbf{Backbone architecture.} We evaluated two different architectures: 3D Sparse Unet [cite] and Point Transformer V3 [cite]. 3D Sparse Unet is an efficient backbone for processing volumetric or point cloud data by leveraging sparse convolutions. The architecture follows the encoder-decoder structure of the traditional U-Net [cite], using downsampling layers to capture high-level features and upsampling layers with skip connections to retain spatial details.

%Point Transformer V3, on the other hand, relies on on the attention mechanism to learn point features. In contrast to its predecessor [cite], Point Transformer V3 is much more efficient due to the serialization to restructure unstructured point clouds and replacing inefficient processes such as K-nearest neighbors (KNN) with efficient serialized neighbor mapping. This design allows the model to learn spatial relationship and features in a more efficient way.

% \textbf{Loss function.} 

%\textbf{Training and monitoring.} 

\subsection{Training and fine-tuning strategy}
\label{sec:training_finetuning_strategy}
We assess the impact of self-supervised pretraining and domain adaptation on downstream performance in data-limited scenarios and compare it to a base scenario: %In particular, we study how models initiated with a pretrained encoder perform compared to those trained from scratch under varying amount of labeled supervision.

\textbf{Training from scratch.} As a baseline, we train the entire model (encoder and task-specific head) using fully labeled data or a subset of it. This setting provides a lower bound for the model performance without leveraging any prior knowledge.

\textbf{Self-supervised pretraining with fine-tuning.} In this setting, we initialize the encoder using weights learned from the contrastive self-supervised learning to capture general geometric and semantic structure. We then fine-tune the model on labeled data from a target region. This strategy allows the model to benefit from large-scale unlabeled forest scene data. 

While self-supervised features provide a strong initialization, they are pretrained in a general and task-agnostic way, and may not be fully aligned with the objectives of specific downstream tasks such as instance segmentation. Consequently, fine-tuning with limited supervision may be insufficient to adapt the model effectively, potentially resulting in suboptimal performance. This motivates us to investigate our next strategy, which introduces domain adaptation to bridge the gap between general-purpose representations and task-specific learning. 

\textbf{Domain adaptation for instance segmentation.} To address the limitations above, we introduce a domain adaptation setup. First, we pretrain the encoder on large-scale unlabeled data. Next, we train the model in a supervised manner on one labeled source to guide the model's learning of task-specific features. Finally, the adapted model is fine-tuned on a target region using only a small fraction of labeled data. This three-stage approach aims to improve transferability across different forest regions while minimizing the need for extensive manual annotation.

In our experiment, domain adaptation is performed using the NIBIO region (which forms part of the FOR-instance dataset) as the labeled source domain and SCION as the target domain. NIBIO represents the boreal forest in Norway, primarily featuring Spruce, and was collected at an extremely high point density. In contrast, SCION, the target domain, is a temperate forest in New Zealand, primarily featuring Pine, and captured at around half the density of NIBIO. 

Even though both are coniferous-dominated forests, the differences in ecological zones, dominant tree species (leading to distinct crown geometries), and data density create a significant domain gap. The pretrained encoder is first adapted via supervised training on NIBIO, followed by fine-tuning on SCION target regions under different proportions of labeled data. 

To prevent the encoder from forgetting the features learned during pretraining, we adopt a two-stage fine-tuning approach. First, we freeze the encoder and train only the head. Then, we unfreeze the entire model and fine-tune all layers. This staged adaptation helps retain general features while enabling the head to adjust precisely to the new domain.

\textbf{Training and fine-tuning for tree classification.} Unlike segmentation tasks, where benchmark data is relatively small, the FOR-species20K offers a large collection of labeled individual trees. To develop a generalizable model for tree classification, we first pretrain a model using supervised learning by merging all available species into two broad categories: broadleaf and coniferous. This grouping allows the model to learn general discriminative features of tree structures without focusing on individual species. 

To evaluate the effectiveness of the pretrained model, we fine-tune it on unseen species using a limited number of samples, simulating a few-shot learning scenario. Specifically, we select four species, e.g., \textit{Pinus sylvestris}, \textit{Picea abies}, \textit{Quercus robur}, and \textit{Acer campestre}. For each species, 1,000 samples are selected, with 40 randomly chosen for training and the remaining used for validation. 

These classes are excluded during pretraining to ensure the model has no prior exposure to them. We evaluate the models under three scenarios: inter-group classification (\textit{Pinus sylvestris} vs. \textit{Quercus robur}) and intra-group classification (\textit{Pinus sylvestris} vs. \textit{Picea abies} and \textit{Quercus robur} vs. \textit{Acer campestre}), thereby assessing both intra- and inter-group separability. For comparison, we also train a model from scratch using the same limited samples. 

\subsection{Label-efficient learning scenarios}
\label{reduction_strategy}

We evaluate two different annotation scenarios for instance segmentation to reduce the labeled data, each strategy reflecting practical constraints faced by human operators in real-world cases.

\textbf{Uniform label reduction.} In this scenario, we reduce the number of labeled points per tree by selecting only a subset of points throughout the scenes. This mimics a situation where a human operator selects sparse labels for each tree, significantly reducing annotation time. Our goal is to identify label quantities feasible for a rapid human annotator, while also examining how model performance behaves as label availability decreases. Specifically, we evaluate six levels of label sparsity: 50\%, 20\%, 10\%, 1\%, 0.1\%, and 0.01\%. 

\begin{figure}[h!]
    \centering
    \begin{subfigure}[b]{0.32\linewidth}
        \includegraphics[width=0.95\linewidth]{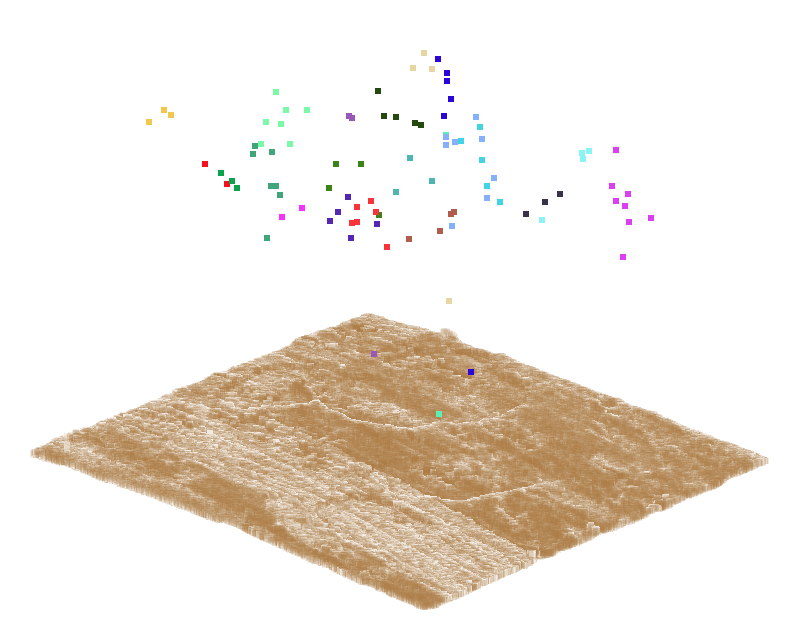}
        \caption{0.01\%}
        \label{fig:site1}
    \end{subfigure}
    \begin{subfigure}[b]{0.32\linewidth}
        \includegraphics[width=0.95\linewidth]{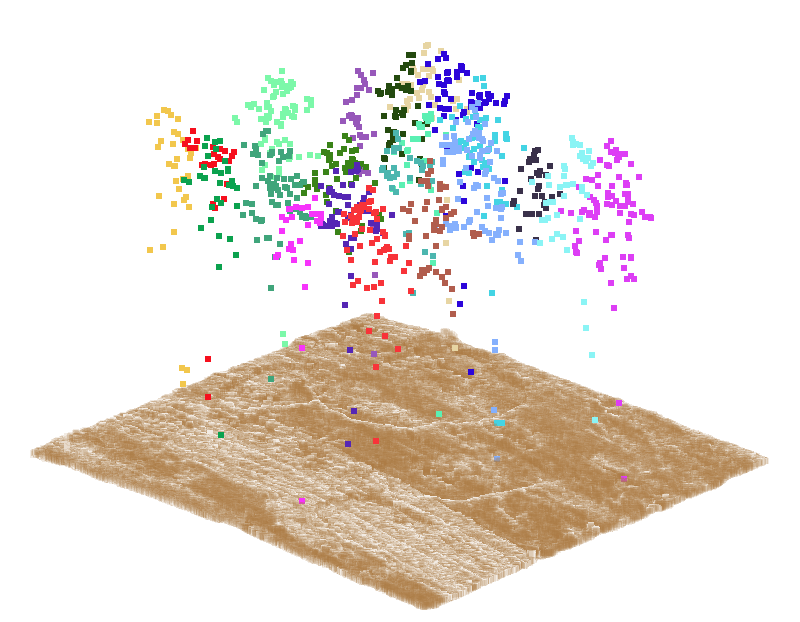}
        \caption{0.1\%}
        \label{fig:site2}
    \end{subfigure}
    \begin{subfigure}[b]{0.32\linewidth}
        \includegraphics[width=0.95\linewidth]{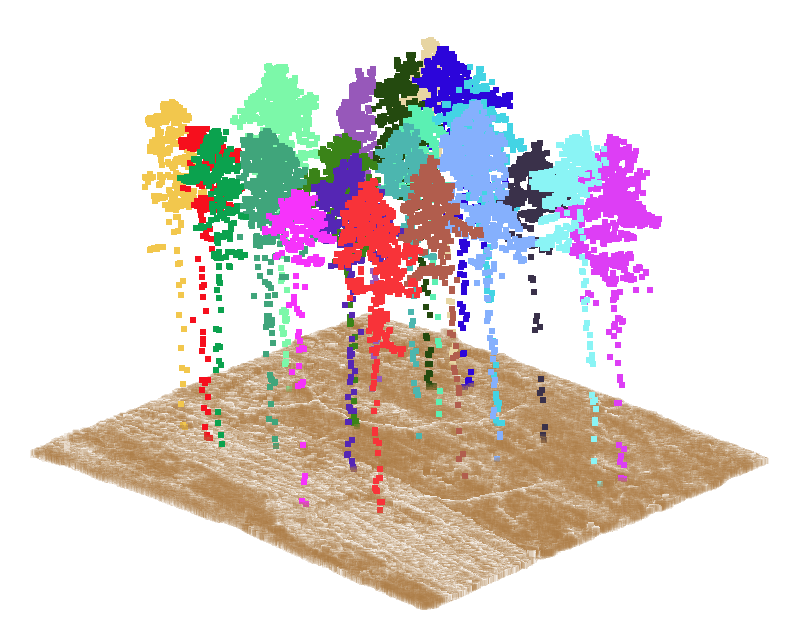}
        \caption{1\%}
        \label{fig:site3}
    \end{subfigure}
    \\
    \begin{subfigure}[b]{0.32\linewidth}
        \includegraphics[width=0.95\linewidth]{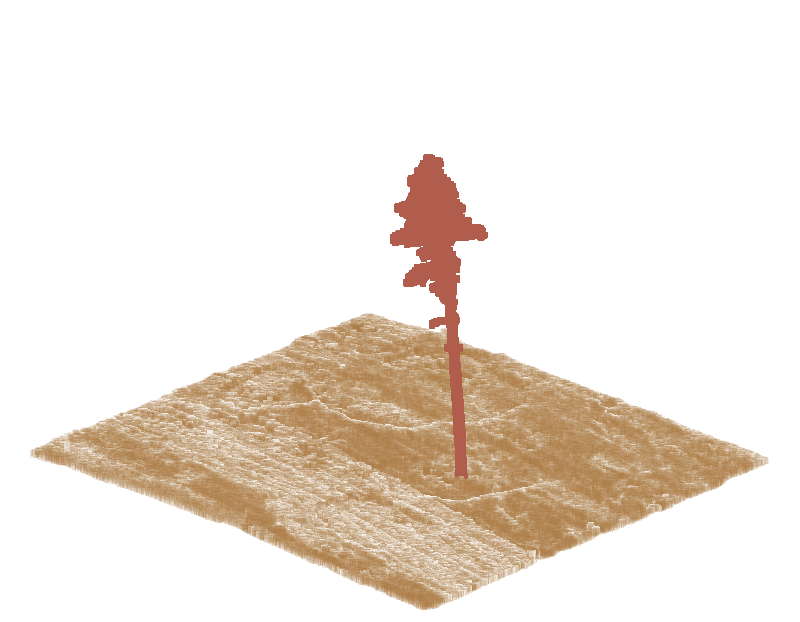}
        \caption{1 tree}
        \label{fig:site4}
    \end{subfigure}
    \begin{subfigure}[b]{0.32\linewidth}
        \includegraphics[width=0.95\linewidth]{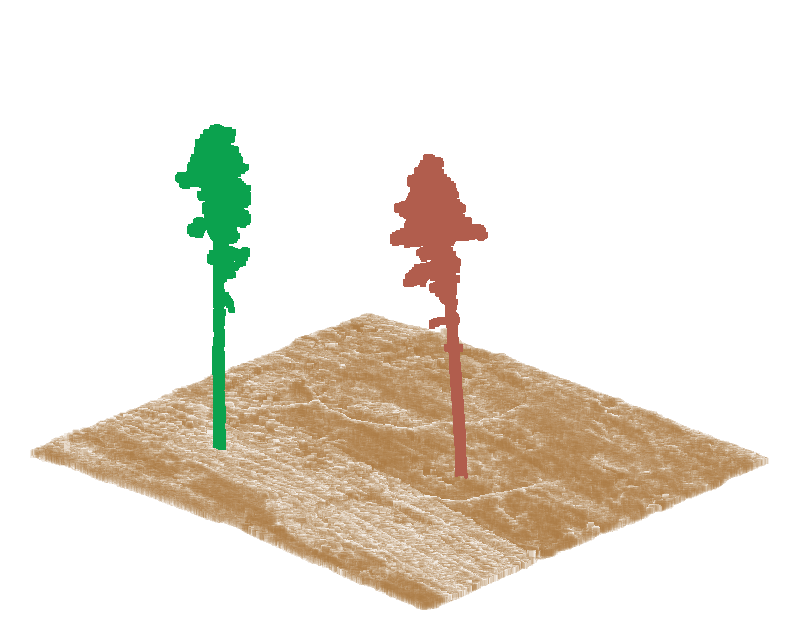}
        \caption{2 trees}
        \label{fig:site5}
    \end{subfigure}
    \begin{subfigure}[b]{0.32\linewidth}
        \includegraphics[width=0.95\linewidth]{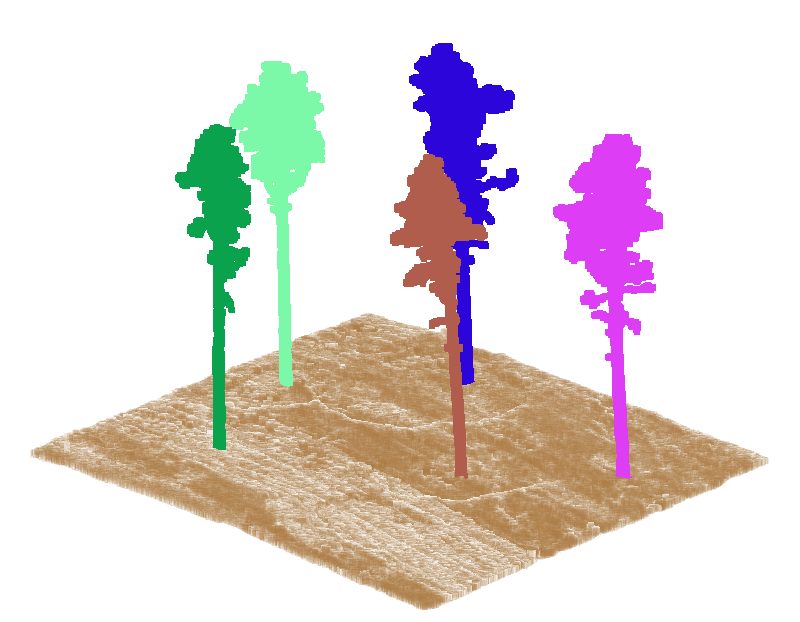}
        \caption{5 trees}
        \label{fig:site6}
    \end{subfigure}
    \\
    \begin{subfigure}[b]{0.32\linewidth}
        \includegraphics[width=0.95\linewidth]{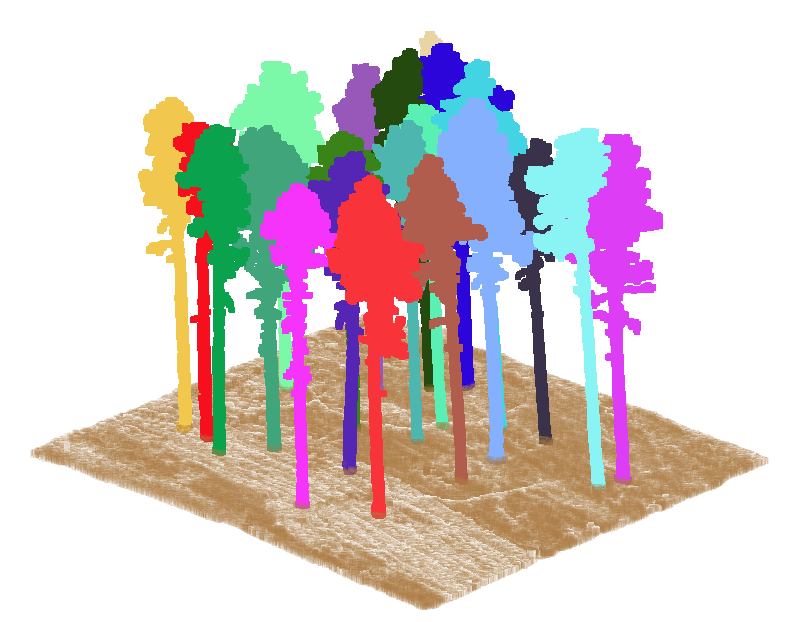}
        \caption{fully labeled points}
        \label{fig:site7}
    \end{subfigure}
    \caption{Examples of sampled points and trees under different labeling scenarios. [a,b,c] illustrate uniform label reduction, while [d,e,f] show tree-level label reduction.}
    \label{fig:label_reduction}
\end{figure}

Table \ref{tab:stats_reduced} shows statistics of the reduced labels per tree in one of the forest scene regions, while Figure \ref{fig:site1}, \ref{fig:site2}, and \ref{fig:site3} show the sparse labels at each reduction level. In a 0.01\% scenario, the average number of points per tree is less than 5 points. Although this level of supervision is extremely sparse and may provide limited information at the instance level, it is included to characterize model behavior at the extreme lower bound of supervision. 

%We evaluate the impact of this sparse supervision using different sampling strategies including:

%\begin{itemize}
    %\item Stratified sampling: to ensure balanced representation of different tree instances or tree components,
    %\item Random sampling: a simple sampling baseline, and
    %\item Farthest Point Sampling (FPS): to maximize spatial coverage of the sampled points. 
%\end{itemize}

\begin{table}[h]
    \centering
    \caption{Statistics of remaining number of points after uniform label reduction}
    \begin{tabular}{c|c}
        \hline
        \textbf{Label proportion} & \textbf{Average} \\
        \hline
        100\% & 49785.5 \\
        50\% & 24892.4 \\ 
        20\% & 9956.8 \\
        10\% & 4978.1 \\
        1\% & 497.3 \\
        0.1\% & 49.3 \\
        0.01\% & 4 \\
        \hline
    \end{tabular}
    \label{tab:stats_reduced}
\end{table}

\textbf{Tree-level label reduction.} Instead of reducing the number of points per tree, this scenario reduces the number of entire annotated trees within each forest scene, as shown in Figure \ref{fig:site4}, \ref{fig:site5}, and \ref{fig:site6}. All points for the selected trees are fully labeled. This approach reflects the practical consideration that annotating a full tree requires a similar amount of effort as annotating only a few points on it. This is particularly valid for instance-level annotation. This strategy reduces labeling effort by limiting the number of annotated trees per scene while preserving full-instance supervision for selected examples. In our experiments, we evaluate this strategy by reducing the number of labeled trees per training scene to 10, 5, 2, and 1. 

\subsection{Branches for Downstream Tasks}
\subsubsection{Instance segmentation branch}

Instance segmentation aims to distinguish individual tree instances within the point cloud, even when they share the same semantic label. This task requires grouping points into unique object instances based on spatial and structural coherence. This is particularly challenging in forest environments due to overlapping crowns, varying densities, and structural similarity between neighboring trees.

In our workflow we follow a center-based approach, where the network learns to predict per-point offset vectors, directing it toward the estimated centroids of its corresponding tree instance. Once the offset vectors are applied to shift the original point coordinates, we apply the Breadth-First Search (BFS) clustering algorithm in the shifted coordinates to group points into distinct tree instances.

\begin{figure}[h!]
    \centering
    \includegraphics[trim= 0 120 0 120,clip,width=0.9\linewidth]{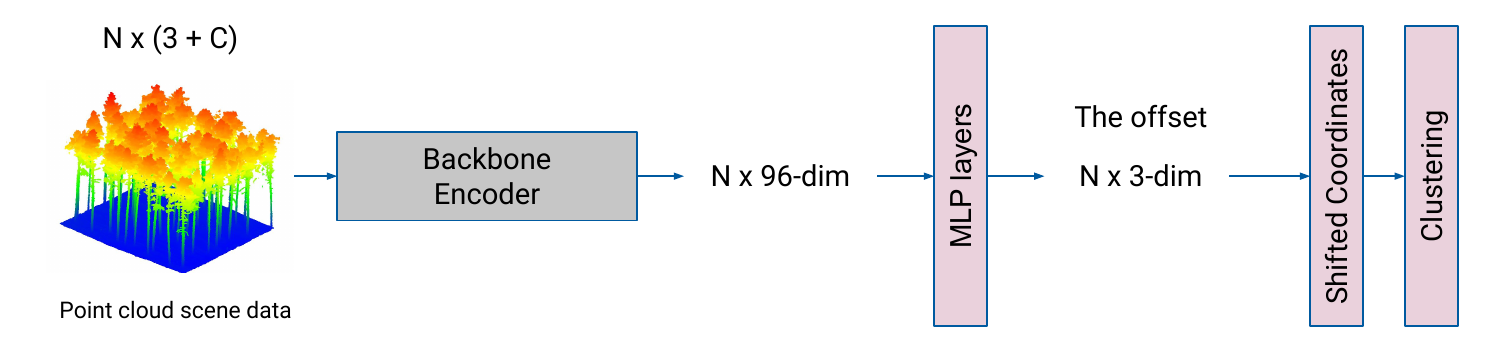}
    \caption{Instance segmentation pipeline. The input consists of $N$ points with 3 spatial coordinates ($X,Y,Z$) and $C$ additional attributes (intensity, return number, and number of returns). The backbone encoder extracts 96-dimensional features used by MLP layers to predict point-wise offsets. Points are then translated to shifted coordinates for final instance clustering.}
    \label{fig:instance_head}
\end{figure}

\textbf{Architecture.} We adopt the PointGroup (\cite{Jiang2020PointGroup}) architecture to perform the instance segmentation task, which builds on the pretrained encoder and adds a lightweight offset head. Specifically, it uses a multilayer perceptron (MLP) composed of two linear layers with an intermediate normalization and ReLU activation. This head predicts a 3D offset vector from each point toward the estimated instance center. Figure \ref{fig:instance_head} illustrates the instance pipeline followed by a clustering.

\textbf{Loss functions.} We supervise the training using a combination of L1 regression loss and cosine similarity loss. L1 loss measures the absolute difference between the predicted offset vectors and the ground truth. The cosine similarity complements this by enforcing directional alignment between the predicted and ground truth vectors, ensuring each point is guided toward its correct instance center. In addition, cross-entropy loss is introduced for semantic learning of tree and non-tree classes. Cross-entropy is chosen to directly compare predicted class probabilities and maximizing the likelihood of the correct class. They are jointly optimized in an end-to-end training for both semantic and instance-level understanding. 

\textbf{Clustering and postprocessing.} After inference, each point is spatially shifted by its predicted offset, effectively repositioning it closer to its estimated instance center. To group points into individual instances, we apply the BFS clustering algorithm in this shifted space by constructing a graph via radius-based neighbor queries. Specifically, points within a defined distance threshold are connected as neighbors, forming clusters through the BFS algorithm that aggregates spatially close points. 

Since our goal is to segment entire trees as single instances, fine-grained semantic distinctions between tree components, such as stems, branches, and leaves, are unnecessary at this stage. To simplify clustering, we remap the detailed semantic predictions into a binary mask distinguishing only between tree and non-tree points. BFS clustering is then performed exclusively on the points labeled as trees in this remapped space. Finally, the resulting instances are assigned their original (pre-remapping) semantic labels, producing panoptic-like outputs in which each segmented tree is associated with a unique instance ID and a fine-grained semantic class.

The following hyperparameters are critical for the reproducibility of our tree segmentation results:

\begin{itemize}
    \item \underline{Distance threshold (\textit{r})}: The radius for neighbor queries in BFS was tuned per dataset to account for varying tree densities. We used a base threshold \textit{r} = 1 for NIBIO and SCION datasets, \textit{r} = 2 for CULS and RMIT, and \textit{r} = 3 for the TUWIEN dataset.
    \item \underline{Cluster filtering}: To eliminate small false-positive groupings and noise, we set a minimum cluster size of 1000 points. Only clusters exceeding this threshold are retained as valid tree instances.
\end{itemize}
    
%\textbf{Training strategy.} We explore both training from scratch and fine-tuning settings. To evaluate data-efficient learning, we train the model using varying percentages of labeled data, randomly sampled from the training set. This allows us to assess how well pretrained features support downstream tasks under limited supervision.

\textbf{Evaluation.} We follow the procedure established by \cite{XIANG2024114078}, utilizing F1-score, Precision, and Recall calculated at a $50\%$ Intersection over Union (IoU) threshold. We also report Commission and Omission Errors, as well as False Positive (FP) and False Negative (FN), to specifically quantify over-detection and under-detection tendencies within the forest canopy. Furthermore, to provide a more granular assessment of model performance across different labeling scenarios, we compute mean Average Precision (mAP) and Average Precision at 50\% IoU (AP50). Especially, mAP provides an overall measure of how well the model detects trees across a range of confidence levels, rather than relying on a single threshold, giving a more complete picture of detection performance.

% \textbf{Evaluation.} We evaluate instance segmentation using various metrics including F1-score, Precision, Recall, Commission and Omission Error, mean Average Precision (mAP), and Average Precision at IoU 50\% (AP50). 

% The metrics used for evaluation include mAP, AP50, F1-score, precision, recall, commission error, omission error, true positive (TP), false positive (FP), and false negative (FN).

\subsubsection{Semantic segmentation branch}

The semantic segmentation branch is designed to assign a class label to each point in the input 3D point cloud. This branch distinguishes between different tree components such as stems, branches, and leaves, as well as low vegetation and terrain. This task is framed as a point-wise classification problem.

\begin{figure}[h!]
    \centering
    \includegraphics[trim= 0 120 0 120,clip,width=0.9\linewidth]{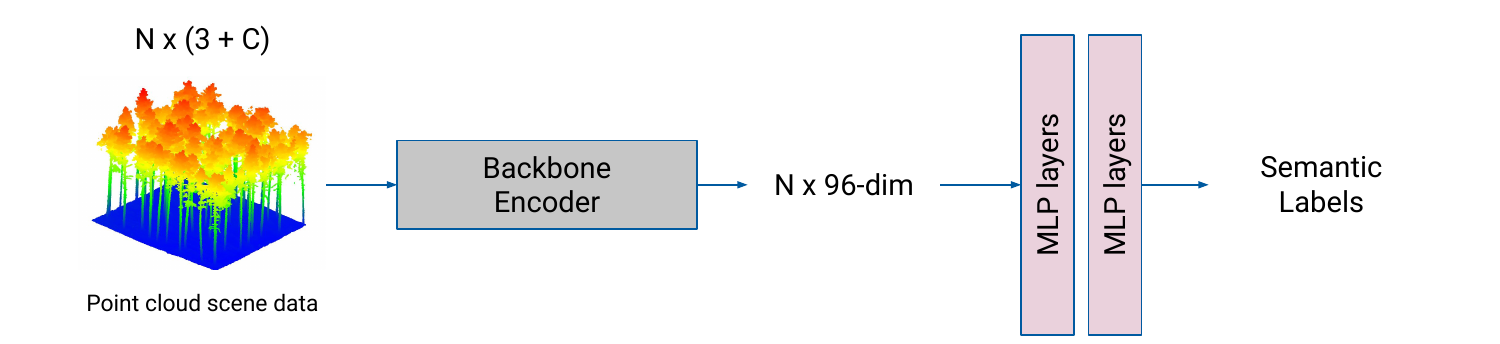}
    \caption{Semantic segmentation pipeline. The input follows the same structure as in instance segmentation pipeline, consisting of $N$ points with 3 spatial coordinates and $C$ additional attributes. The backbone encoder extracts features, which are processed by dual MLP layers to predict point-wise semantic labels.}
    \label{fig:semantic_head}
\end{figure}

\textbf{Architecture.} We adopt Sparse UNet architecture to perform this task and build the semantic branch upon the pretrained encoder. The main reason for selecting Sparse UNet is to maintain architectural continuity with the pretraining phase. The pretrained encoder serves as weights initialization. The encoder outputs contextualized per-point embeddings. On top of this encoder, we attach a segmentation head consisting of a lightweight MLP with shared weights across all points. This head projects the feature vectors to logits, which can then be converted into class probabilities. Figure \ref{fig:semantic_head} shows the semantic head with the MLP layers.

\textbf{Loss function.} We use a cross-entropy loss for structural classes of tree in a point-wise semantic classification and apply inverse class frequency weighting to handle class imbalance.

%\textbf{Training strategy.} Similar to instance segmentation task, we explore two learning setups: training from scratch and fine-tuning from pretrained weights. We also evaluate data-efficient learning by randomly reducing labeled data from the training set.

%\textbf{Evaluation.} Semantic segmentation performance is evaluated using mean Intersection over Union (mIoU), Overall Accuracy (OA), per-class accuracy, and F1-score. 

\textbf{Evaluation.} We utilize IoU and Accuracy, calculated for each individual class, alongside mean IoU and mean Accuracy to account for class imbalance. These metrics provide a balanced average across all classes, ensuring that the performance on rare classes is weighted equally with dominant ones.

% We report results under different supervision levels to asses benefits of pretraining. Additionally, we employ traditional machine learning baselines, such as random forests and support vector machines trained on handcrafted geometric features, to contextualize the performance gains of deep, pretrained representations. This comparison highlights the strengths of learned features in capturing complex spatial and semantic patterns in forest scenes.

\subsubsection{Individual tree classification branch}

The goal of the individual tree classification branch is to assign a semantic class label to each individual tree instance, such as species or tree type. 
% This branch operates after instance segmentation, where individual tree point clouds have already been isolated.

%\textbf{Instance extraction.} We use the predicted instance masks from the instance segmentation branch to extract point clouds corresponding to individual trees. Each tree instance is represented as a set of 3D points with associated features derived from the pretrained backbone:

%\begin{equation}
    %I_k = %\begin{Bmatrix}\textbf{p}_i,\textbf{f}_i\end{Bmatrix}_{i=1}^{N_k}
%\end{equation}

%where $N_k$ is the number of points in the instance, $\textbf{p}_i$ is the 3D coordinate, and $\textbf{f}_i$ is the corresponding feature vector from the encoder.

\begin{figure}[h]
    \centering
    \includegraphics[trim= 0 80 0 80,clip,width=0.9\linewidth]{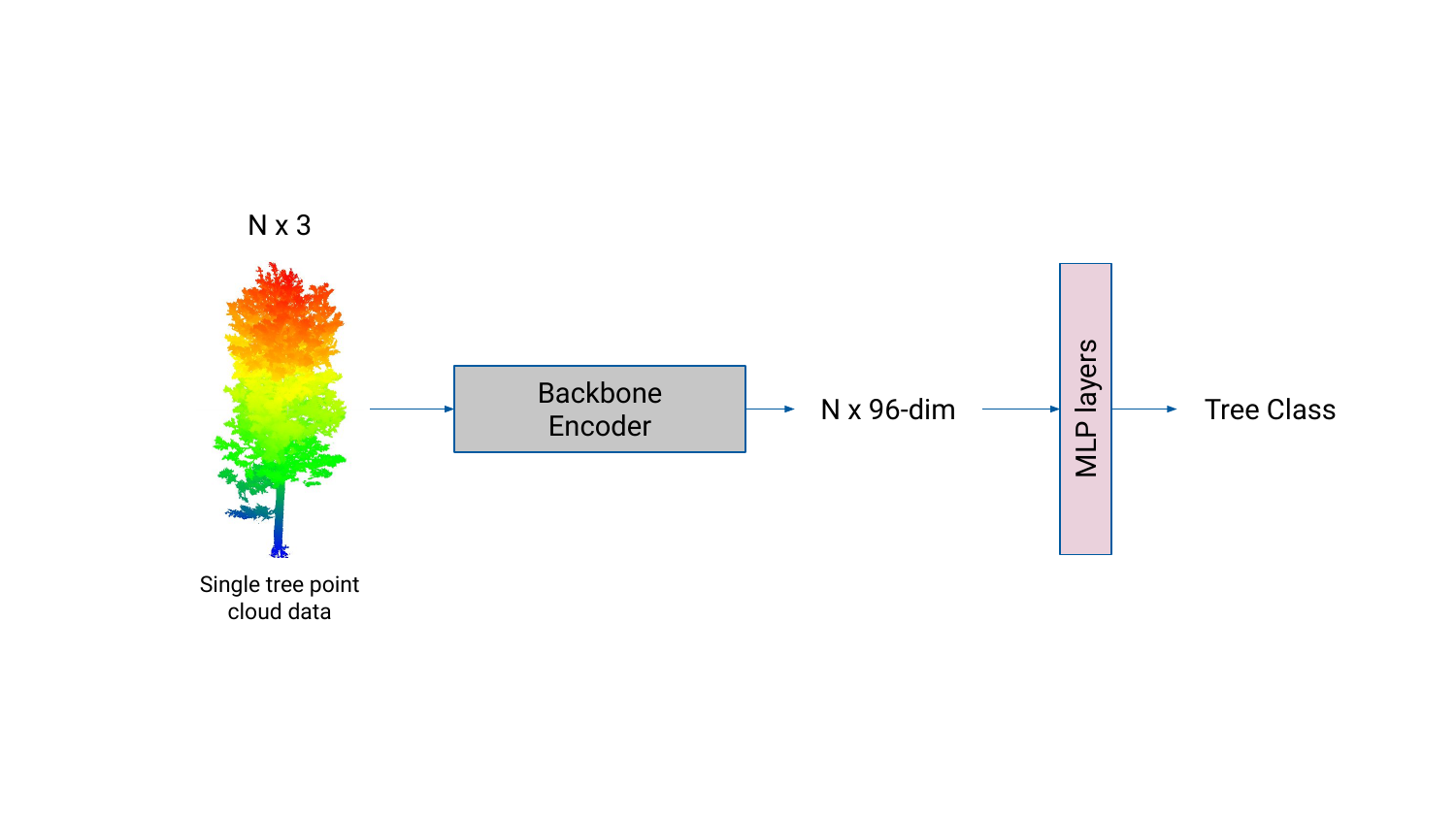}
    \caption{Tree classification pipeline. Features from the backbone encoder are aggregated to predict final tree class.}
    \label{fig:classification_head}
\end{figure}

\textbf{Architecture.} Motivated by its high computational efficiency on the inherently sparse data of point clouds, we employ a SparseUnet model to classify tree instances. The design aggregates local features into a global representation. It works by taking per-point features from the pretrained encoder and feeding them through small, shared MLPs across all points in the instance to generate a fixed-size global feature vector. A softmax classifier then predicts the tree class label. Figure \ref{fig:classification_head} illustrates the classification head.

\textbf{Loss function.} The classifier is trained using a cross-entropy loss over the predicted class probabilities and ground truth labels. Cross-entropy loss is a standard for multi-class classification because it minimizes the distance between the predicted probability and the true distribution.

%\textbf{Evaluation.} We evaluate individual tree classification using Overall Accuracy (OA), per-class precision, recall, and F1-score. In addition, we use confusion matrix to analyze misclassifications patterns. 

\textbf{Evaluation.} We utilize Jaccard index and Accuracy, computed for each class individually, alongside mean Jaccard and mean Accuracy to provide an aggregated assessment.

% To evaluate the effectiveness of data-efficient learning, we reduce the number of training samples, where each sample corresponds to one segmented tree instance (i.e., one point cloud of a single tree). %This is a different strategy from semantic and instance segmentation, where we reduced the proportion of labeled points within full-scene level point clouds.

\subsection{Implementation details}
We pretrain the backbone network using the Stochastic Gradient Descent (SGD) optimizer with Nesterov acceleration, which is known to find flatter minima for better model generalization than adaptive methods like Adam. We set the momentum to 0.8 to ensure stable movement through the loss landscape and use a standard weight decay of $1e-4$ to prevent overfitting. We train for 1200 epochs using the One Cycle learning rate scheduler, for fast convergence by cycling the learning rate between a low and a high bound (\cite{smith2019super}), with:

\begin{itemize}
    \item a maximum learning rate of 0.1,
    \item warm-up phase covering the first 1\% of training,
    \item cosine annealing as the decay strategy,
    \item initial learning rate = 0.01, and
    \item final learning rate $\approx 1e-6$.
\end{itemize}

For the downstream tasks, we fine-tune the pretrained backbone using SGD optimizer with Nesterov acceleration, a momentum of 0.9, and a weight decay of $1e-4$. However, we switch to a polynomial learning rate scheduler with an initial learning rate of 0.1, training for 3000 epochs. The polynomial learning rate is preferred for fine-tuning because its smooth, gradual decay the learning rate prevents large, destabilizing updates to the pretrained weights (\cite{mishra2019polynomial}), thus mitigating the risk of catastrophe forgetting and allowing for gentle adaptation for task-specific data. The whole pipeline is implemented using the Pointcept codebase (\url{https://github.com/Pointcept/Pointcept}).

%%%%%%%%%%%%%%%%%%%%%%%%%%%%%%%%%%%%%
\section{Results}
\label{sec:results}

\subsection{Contrastive pretraining}

To validate the effectiveness of our self-supervised contrastive pretraining, we analyze the learning dynamics using two key metrics: contrastive InfoNCE loss and cosine similarities. The InfoNCE loss describes the log-likelihood of correctly identifying the positive sample among a set of negative samples, while cosine similarities describe that the mean similarity of positive pairs should increase toward 1.0 and negative pairs should decrease toward 0.0. 

\begin{figure}[h]
    \centering
    \includegraphics[width=0.8\linewidth]{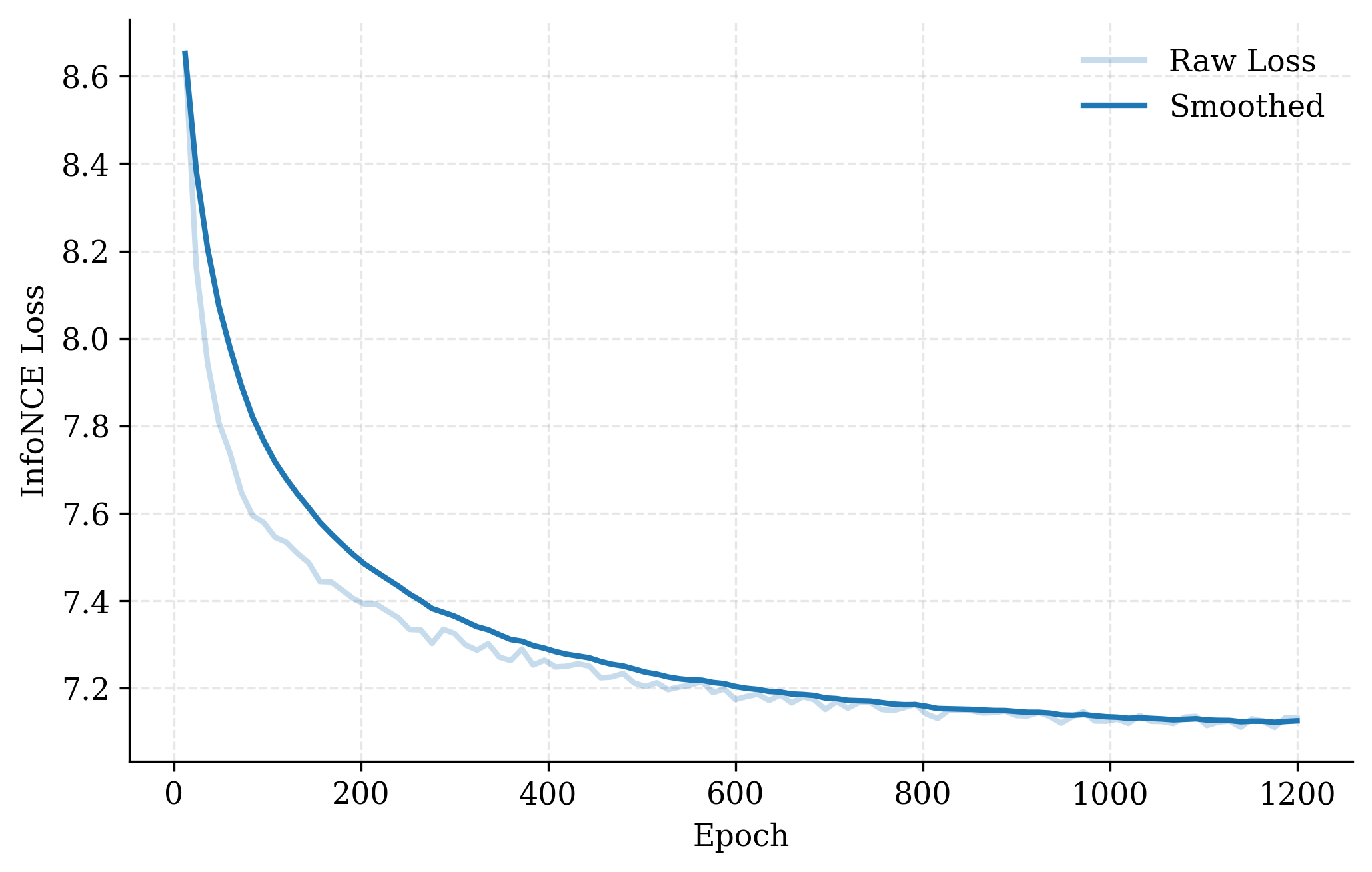}
    \caption{Loss monitoring during contrastive pretraining.}
    \label{fig:contrastive_pretraining}
\end{figure}

\textbf{Contrastive InfoNCE loss curve.} We observe a steady decrease in contrastive loss over the course of training, as shown in Figure \ref{fig:contrastive_pretraining}, indicating successful representation learning. The loss converges smoothly, suggesting stable optimization and appropriate hyperparameters. 

\begin{figure}[h]
    \centering
    \begin{subfigure}[b]{0.45\textwidth}
        \centering
        \includegraphics[width=\textwidth]{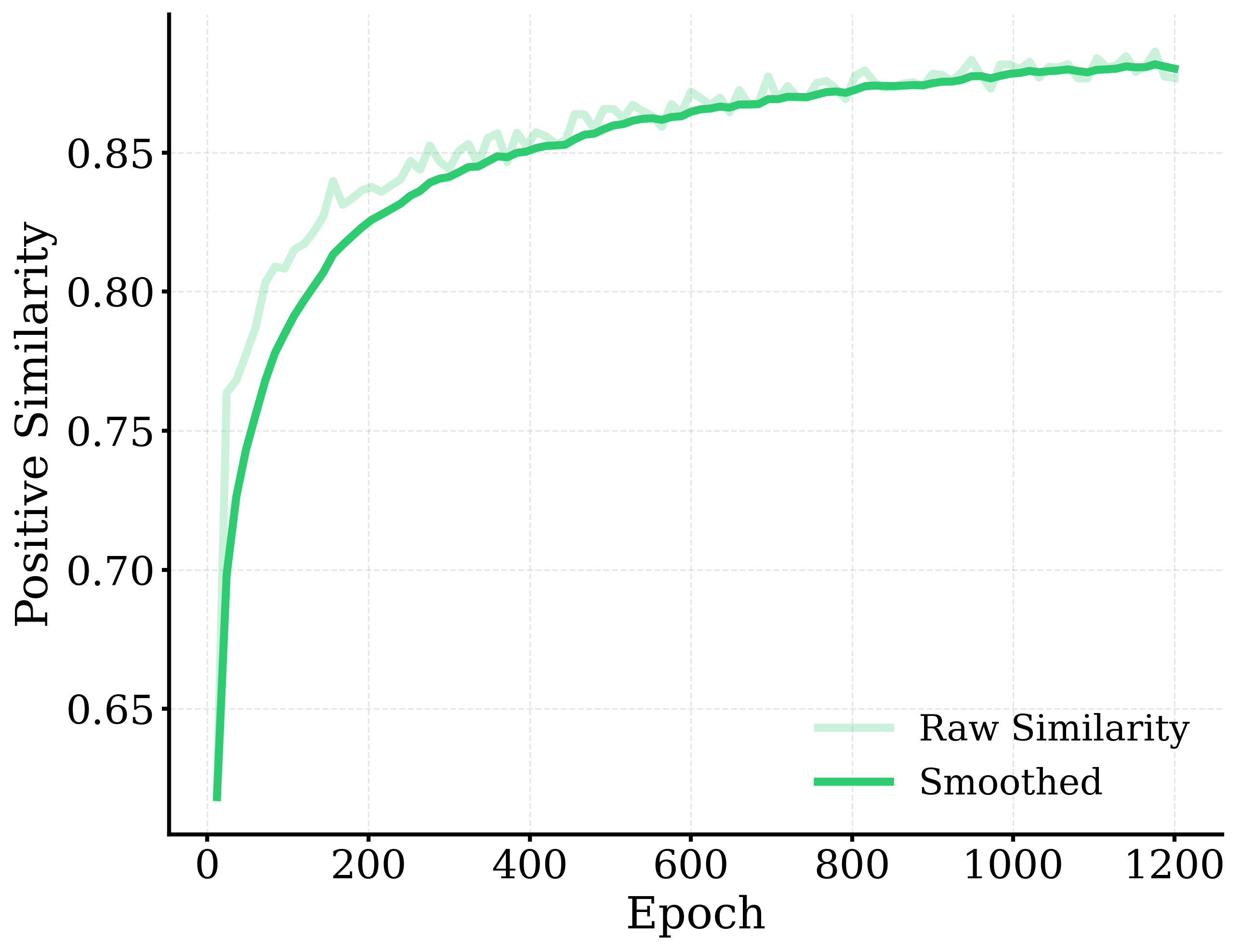}
        \caption{Positive similarity}
    \end{subfigure}
    \hfill
    \begin{subfigure}[b]{0.45\textwidth}
        \centering
        \includegraphics[width=\textwidth]{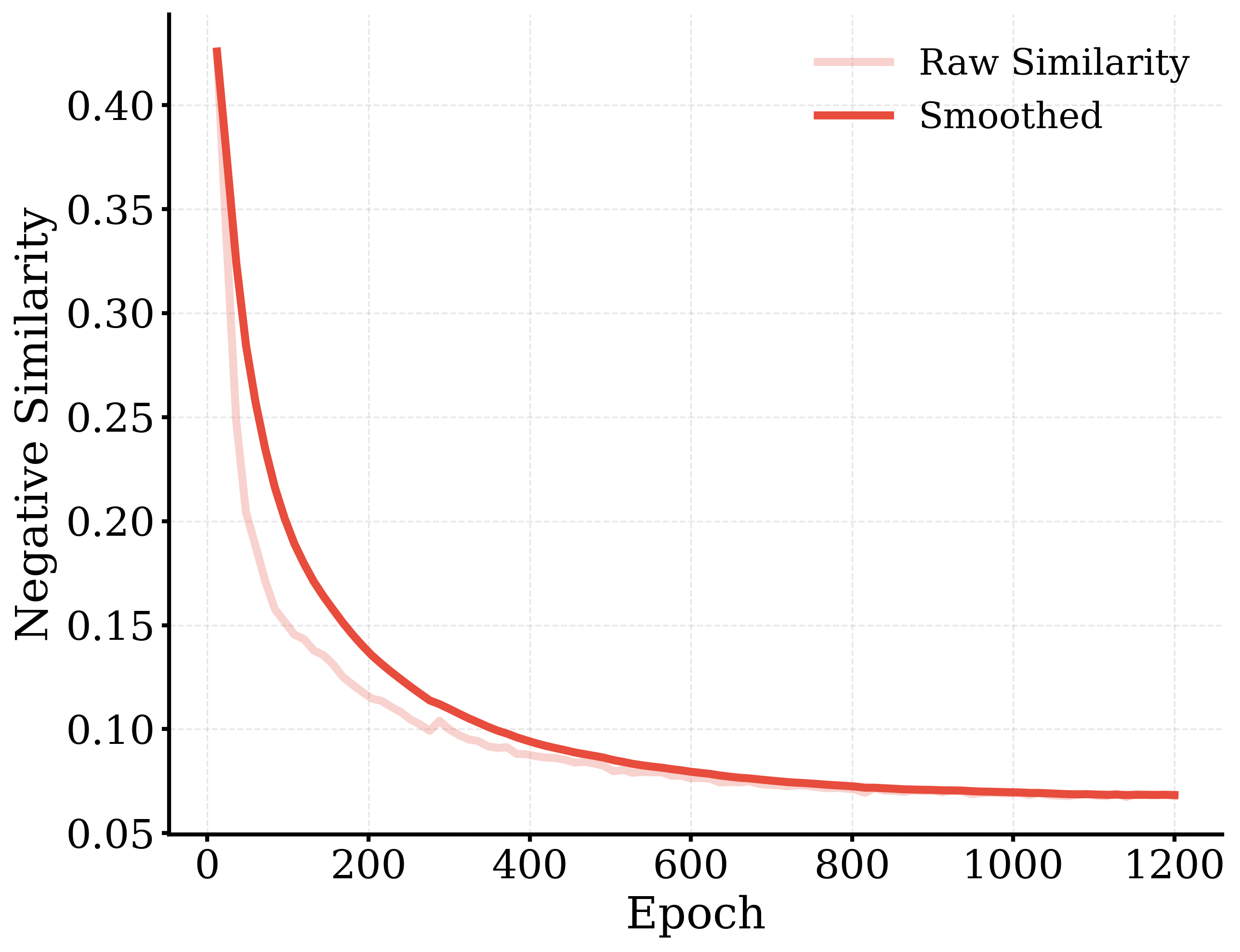}
        \caption{Negative similarity}
    \end{subfigure}
    \caption{Similarity metrics during contrastive pretraining. (a) Positive similarity shows the average cosine similarity between augmented views of the same point cloud sample. (b) Negative similarity shows the average cosine similarity between different point cloud samples.}
    \label{fig:contrastive_pretraining_pos_neg}
\end{figure}

\textbf{Positive vs negative pair similarities.} We track the cosine similarity between anchor-positive and anchor-negative pairs throughout the self-supervised pretraining phase, as illustrated in Figure \ref{fig:contrastive_pretraining_pos_neg}. As training progresses, positive pair similarities increase, approaching 1.0. In contrast, the similarity of negative pairs decreases, typically approaching 0.0, suggesting that the model successfully separates dissimilar features in the representation space.

This widening gap between positive and negative similarities is a strong indication that the contrastive objective is working effectively, leading to more discriminant and meaningful feature embeddings.

% \textbf{Influence of the data augmentation.} [[Here we show how different data augmentation methods affect the results]]

\subsection{Instance segmentation results}

\textbf{Full supervision.} We use the FOR-instance dataset to benchmark the instance segmentation performance across various forest types included in the benchmark dataset. We follow the training and evaluation split as described in the dataset documentation (\cite{puliti_2023_8287792}). Specifically, we utilize the designated 21 development plots for model training and hyperparameter tuning (using a 16/5 split for internal validation), and the 11 test plots for final evaluation. 

\begin{figure*}[ht!]
    \centering
    \includegraphics[width=0.9\linewidth]{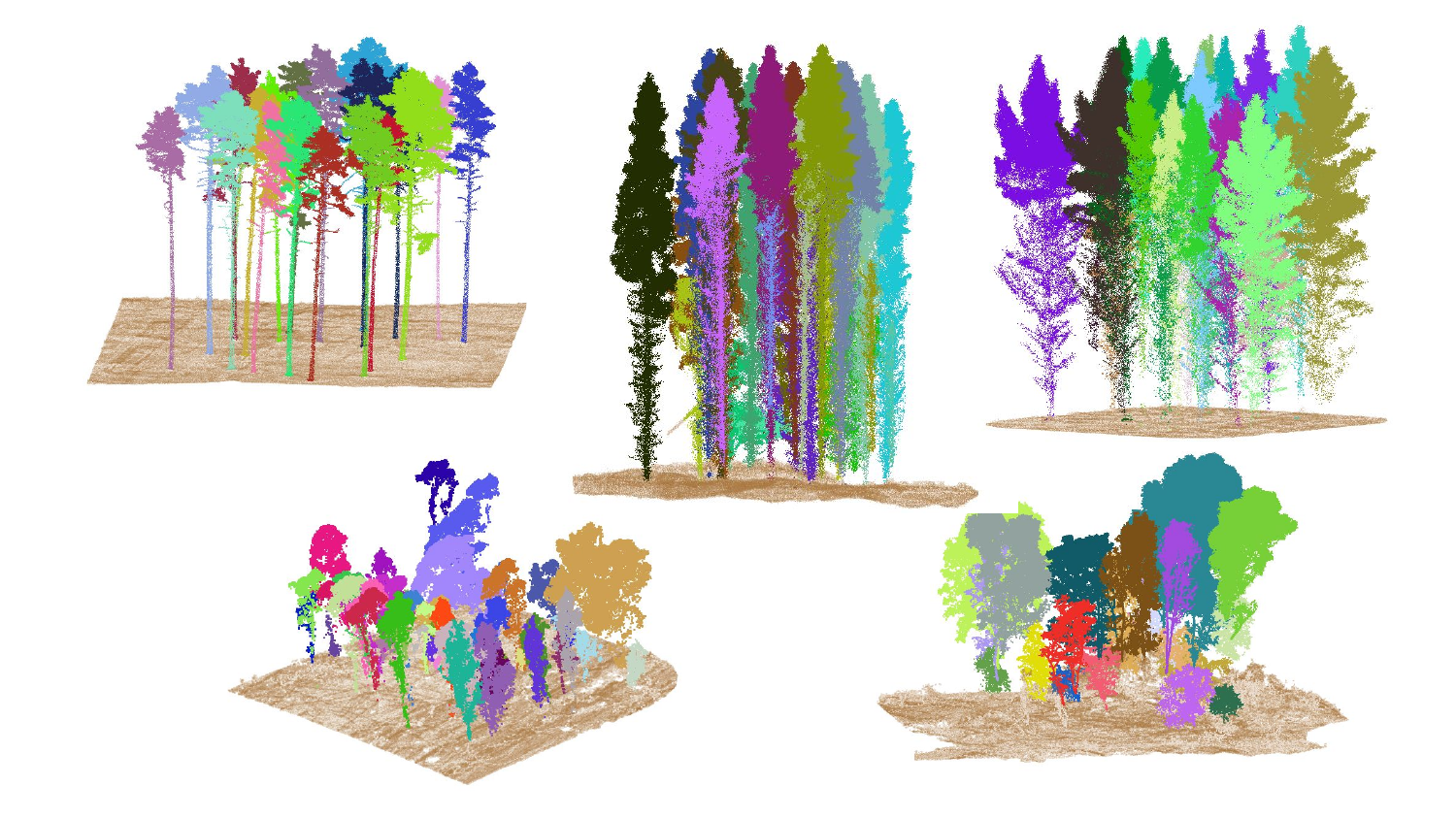}
    \caption{Visualization of instance segmentation  results on the different forest regions in the FOR-instance dataset.}
    \label{fig:all_insseg_results}
\end{figure*}

Table \ref{tab:insseg_all} summarizes the results across different forest types, reported under the fully supervised setting to provide an upper-bound reference for model performance. Finally, Figure \ref{fig:all_insseg_results} shows the prediction results of various scenes within the FOR-instance dataset.

%The metrics used for evaluation include mAP, AP50, F1-score, precision, recall, commission error, omission error, true positive (TP), false positive (FP), and false negative (FN). These results are reported under the fully supervised setting, which provides an upper-bound reference for model performance.

\begin{table*}[ht!]
    \centering
    \caption{Instance segmentation performance across different forest types}
    \resizebox{\textwidth}{!}{%
    \begin{tabular}{c|c|c|c|c|c|c|c|c}
        \hline
        \multirow{2}{*}{\textbf{Forest types}} & \textbf{F1} & \textbf{Precision} & \textbf{Recall} & \textbf{Commiss.} & \textbf{Omiss.} & \textbf{TP} & \textbf{FP}& \textbf{FN} \\
         & (\%) $\uparrow$ & (\%) $\uparrow$& (\%) $\uparrow$& \textbf{Err.} (\%) $\downarrow$& \textbf{Err.} (\%) $\downarrow$& $\uparrow$ & $\downarrow$& $\downarrow$ \\
        \hline
        CULS\_2  & 100.00 & 100.00 & 100.00 & 0.00 & 0.00 & 20 & 0 & 0 \\
        NIBIO\_1  & 86.15  & 100.00 & 75.67  & 0.00 & 0.24 & 28 & 0 & 9 \\
        NIBIO\_5  & 89.47  & 89.47  & 89.47  & 0.11 & 0.11 & 17 & 2 & 2 \\
        NIBIO\_17  & 92.85  & 100.00 & 86.67  & 0.00 & 0.13 & 26 & 0 & 4 \\
        NIBIO\_18  & 96.15  & 100.00 & 92.59  & 0.00 & 0.07 & 25 & 0 & 2 \\
        NIBIO\_22  & 91.89  & 100.00 & 85.00  & 0.00 & 0.15 & 17 & 0 & 3 \\
        NIBIO\_23  & 90.19  & 100.00 & 82.14  & 0.00 & 0.18 & 23 & 0 & 5 \\
        RMIT      & 70.90  & 84.78  & 60.93  & 0.15 & 0.39 & 39 & 7 & 25 \\
        SCION\_31  & 93.61  & 100.00 & 88.00  & 0.00 & 0.12 & 22 & 0 & 3 \\
        SCION\_61  & 87.50  & 100.00 & 77.77  & 0.00 & 0.22 & 14 & 0 & 4 \\
        TU-WIEN    & 54.54  & 75.00  & 42.85  & 0.25 & 0.57 & 15 & 5 & 20 \\
        \hline
    \end{tabular}}
    \label{tab:insseg_all}
\end{table*}

Our full supervision evaluation demonstrates strong instance segmentation performance across most forest sites, with high accuracy and reliable instance detection metrics. Notably, several sites such as CULS, NIBIO, and SCION achieve perfect precision and zero false positives, reflecting the model’s robustness in those environments. However, performance varies with site complexity. For instance, RMIT and TU-WIEN exhibit lower recall and higher omission errors, indicating challenges in segmenting more heterogeneous or densely forested areas. %These variations suggest that forest structural diversity significantly impacts model effectiveness and point to potential areas for architectural or data-driven improvements.

\textbf{Limited supervision.} To assess the effectiveness of pretraining, we compare models initialized with pretrained weights and domain adaptation against models trained from scratch under various levels of labeled training data reduction using two different label reduction strategies: uniform and tree-level label reduction. We report both full and head-only fine-tuning to understand further how different adaptation strategies perform.

The SCION dataset was selected as the target domain for these experiments to provide a rigorous test of global transferability. As detailed in Section \ref{sec:training_finetuning_strategy}, the significant geographic and structural disparity between the source (NIBIO, Norway) and target (SCION, New Zealand) environments, characterized by different species and point density, serves as a stress test for the framework's ability to generalize to new forest types with minimal supervision.

\textit{\underline{Uniform label reduction.}} We reduce the number of labeled points per tree using stratified sampling and evaluate label proportions of 100\%, 50\%, 20\%, 10\%, 1\%, 0.1\%, and 0.01\%. Table \ref{tab:insseg_uniform_scion} shows the instance segmentation performance (mAP and AP50) on the SCION forest region. 

\begin{table*}[ht!]
    \centering
    \caption{Instance segmentation performance (mAP and AP50, \%) under uniform label reduction on the SCION dataset.}
    \label{tab:insseg_uniform_scion}
    \resizebox{\textwidth}{!}{%
    \begin{tabular}{c|cc|cc|cc}
        \hline
        \multicolumn{7}{c}{\textbf{Uniform Label Reduction}}\\
        \hline
        \multirow{2}{*}{\textbf{Labeled points}}
        & \multicolumn{2}{c|}{\textbf{Fine-tune Head}} 
        & \multicolumn{2}{c|}{\textbf{Fine-tune All}} 
        & \multicolumn{2}{c}{\textbf{From Scratch}} \\
        \cline{2-7}
        & mAP (\%) & AP50 (\%) 
        & mAP (\%) & AP50 (\%) 
        & mAP (\%) & AP50 (\%) \\
        \hline
        100\%  & 52.06 & 67.58 & 76.72 & 89.36 & 43.19 & 72.38 \\
        50\%   & 52.64 & 68.85 & 78.55 & 90.64 & 44.70 & 67.32 \\
        20\%   & 52.80 & 68.16 & 76.26 & 87.78 & 40.10 & 68.96 \\
        10\%   & 52.57 & 68.71 & 77.40 & 88.02 & 46.78 & 74.02 \\
        1\%    & 53.10 & 68.99 & 76.77 & 87.68 & 45.56 & 73.47 \\
        0.1\%  & 54.10 & 67.95 & 66.87 & 84.62 & 33.79 & 65.33 \\
        0.01\% & 38.96 & 71.09 & 21.34 & 48.68 & 0.00  & 0.00 \\
        \hline
    \end{tabular}}
    \vspace{2mm}
\end{table*}

% Interestingly, the performance remains relatively stable when reducing labeled data from 100\% to 1\%, regardless of the fine-tuning strategy or whether training is done from scratch. 

Even though the number of points per tree is reduced gradually, the performance remains relatively stable when moving from 100\% to 1\% labeled data. This is expected because even at the 1\% level, there are still sufficient labeled points per tree to effectively represent the individual instances. However, performance drops significantly under extremely low scarcity, such as 0.1\% and 0.01\%, where the points density become too sparse. While both full fine-tuning and from-scratch training struggle to maintain accuracy, the model fine-tuned using pretrained weights remains superior. 
% This suggests that instance segmentation does not require dense labels per tree, but sparse sampled points per tree are sufficient for the model to learn reliable offset predictions. 

Interestingly, the 50\% label setting achieves the highest performance under full fine-tuning. We attribute this to the high redundancy in dense point clouds and the voxelization in the Sparse UNet backbone (5 cm), where multiple points often fall into the same voxel. Subsampling to 50\% largely preserves the effective voxelized representation while slightly reducing redundancy and ambiguity in overlapping regions, leading to marginally improved instance separation.

\textit{\underline{Tree-level label reduction.}} We evaluate instance segmentation performance by selecting 10, 5, 2, and 1 fully labeled trees per forest scene, using all points within those trees as labeled data. The models are tested under full fine-tuning, head-only fine-tuning, and from-scratch training. 

Table \ref{tab:insseg_treelevel_scion} reports mAP and AP50 scores across these settings. Unlike the uniform label reduction experiment, both full-fine-tuning and from-scratch training immediately exhibit a marked performance decline as the number of labeled trees decreases. Interestingly, head-only fine-tuning consistently yields better generalization, once the number of available labeled trees is less than 5 trees per scene. 

\begin{table*}[hb!]
    \centering
    \caption{Instance segmentation performance (mAP and AP50, \%) under tree-level label reduction on the SCION dataset.}
    \label{tab:insseg_treelevel_scion}
    \resizebox{\textwidth}{!}{%
    \begin{tabular}{c|cc|cc|cc}
        \hline
        \multicolumn{7}{c}{\textbf{Tree-level Label Reduction}}\\
        \hline
        \multirow{2}{*}{\textbf{Labeled trees}} 
        & \multicolumn{2}{c|}{\textbf{Fine-tune Head}} 
        & \multicolumn{2}{c|}{\textbf{Fine-tune All}} 
        & \multicolumn{2}{c}{\textbf{From Scratch}} \\
        \cline{2-7}
        & mAP (\%) & AP50 (\%) 
        & mAP (\%) & AP50 (\%) 
        & mAP (\%) & AP50 (\%) \\
        \hline
        All-tree & 52.06 & 67.58 & 76.72 & 89.36 & 43.19 & 72.38 \\
        10-tree  & 51.46 & 68.14 & 58.65 & 80.04 & 41.79 & 72.62 \\
        5-tree   & 53.49 & 72.02 & 52.32 & 71.97 & 25.22 & 53.55 \\
        2-tree   & 50.84 & 68.49 & 39.89 & 60.97 & 14.33 & 44.53 \\
        1-tree   & 54.17 & 72.75 & 16.23 & 41.63 & 1.05  & 3.93 \\
        \hline
    \end{tabular}}
    \vspace{2mm}
\end{table*}

\begin{figure*}[ht!]
     \centering
     \begin{subfigure}[b]{0.45\textwidth}
         \centering
         \includegraphics[width=\textwidth]{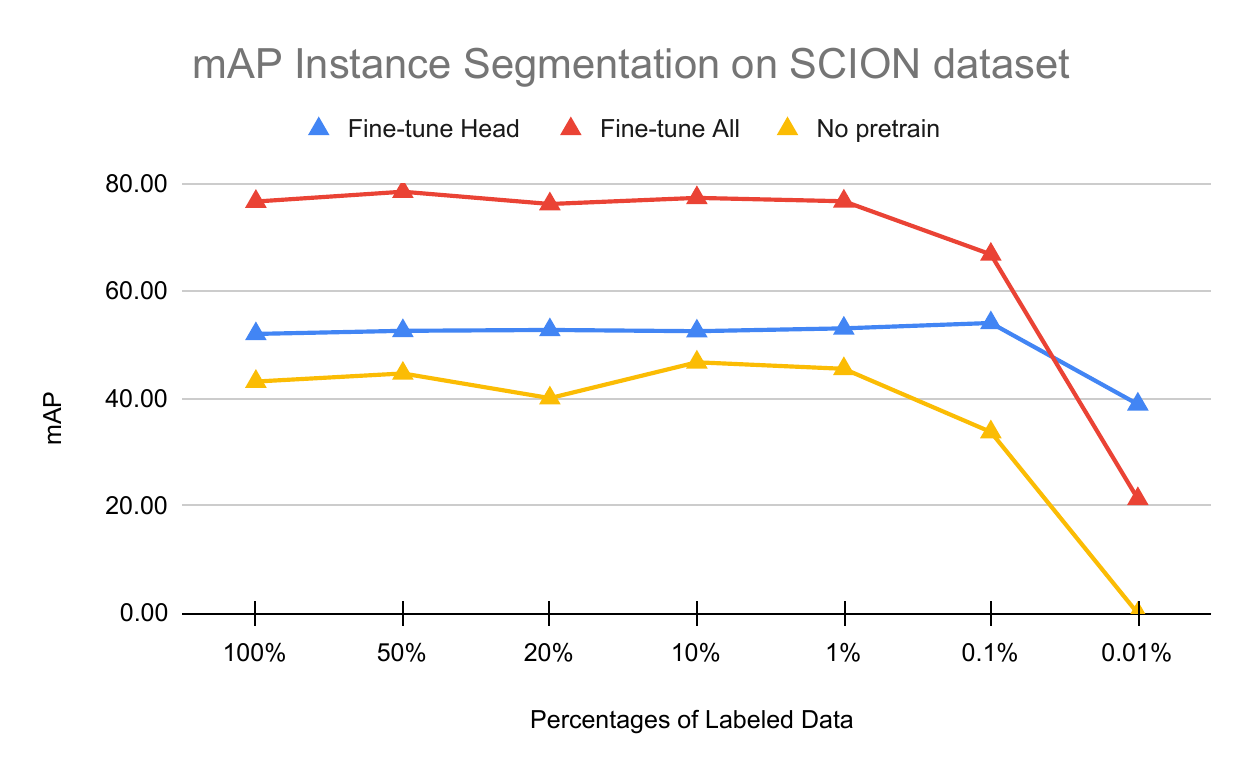}
         \caption{}
     \end{subfigure}
     \begin{subfigure}[b]{0.45\textwidth}
         \centering
         \includegraphics[width=\textwidth]{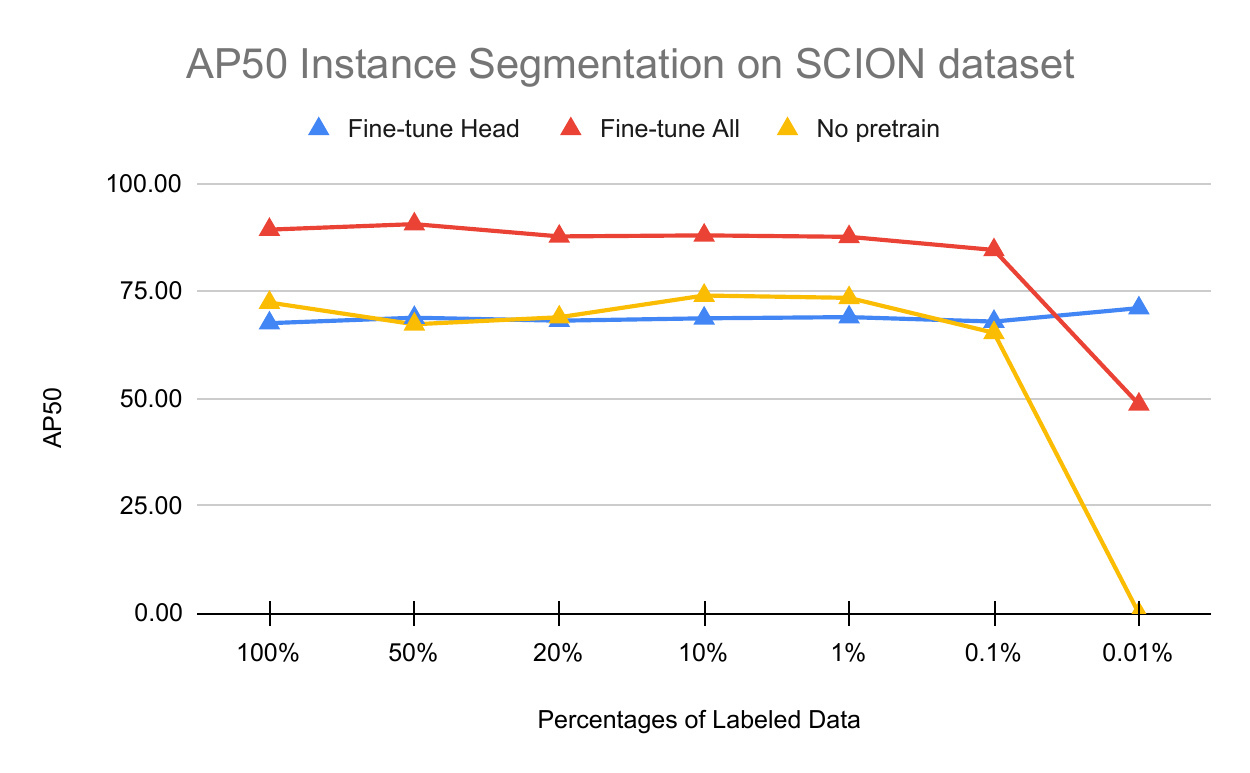}
         \caption{}
     \end{subfigure}
     \caption{Performance of instance segmentation under uniform label reduction using stratified sampling on the SCION dataset.}
     \label{fig:insseg_uniform}
\end{figure*}

\begin{figure*}[ht!]
     \centering
     \begin{subfigure}[b]{0.45\textwidth}
         \centering
         \includegraphics[width=\textwidth]{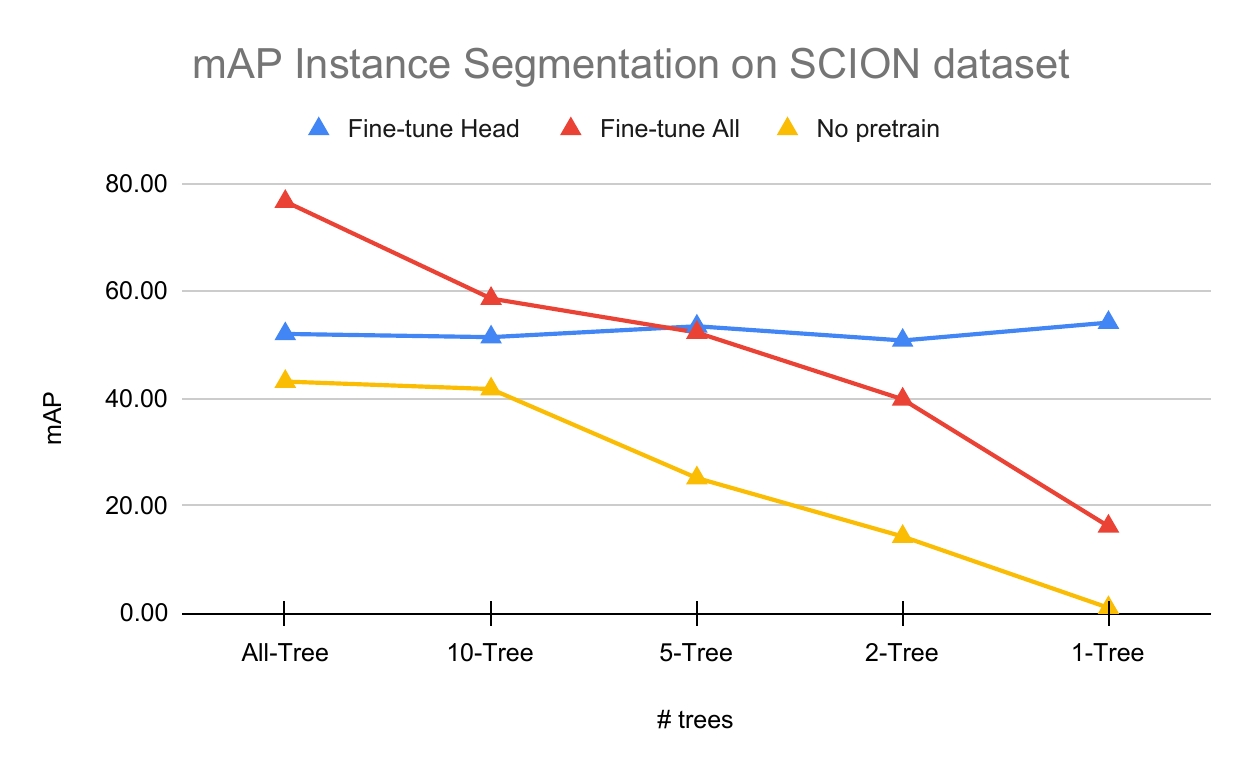}
         \caption{}
     \end{subfigure}
     \begin{subfigure}[b]{0.45\textwidth}
         \centering
         \includegraphics[width=\textwidth]{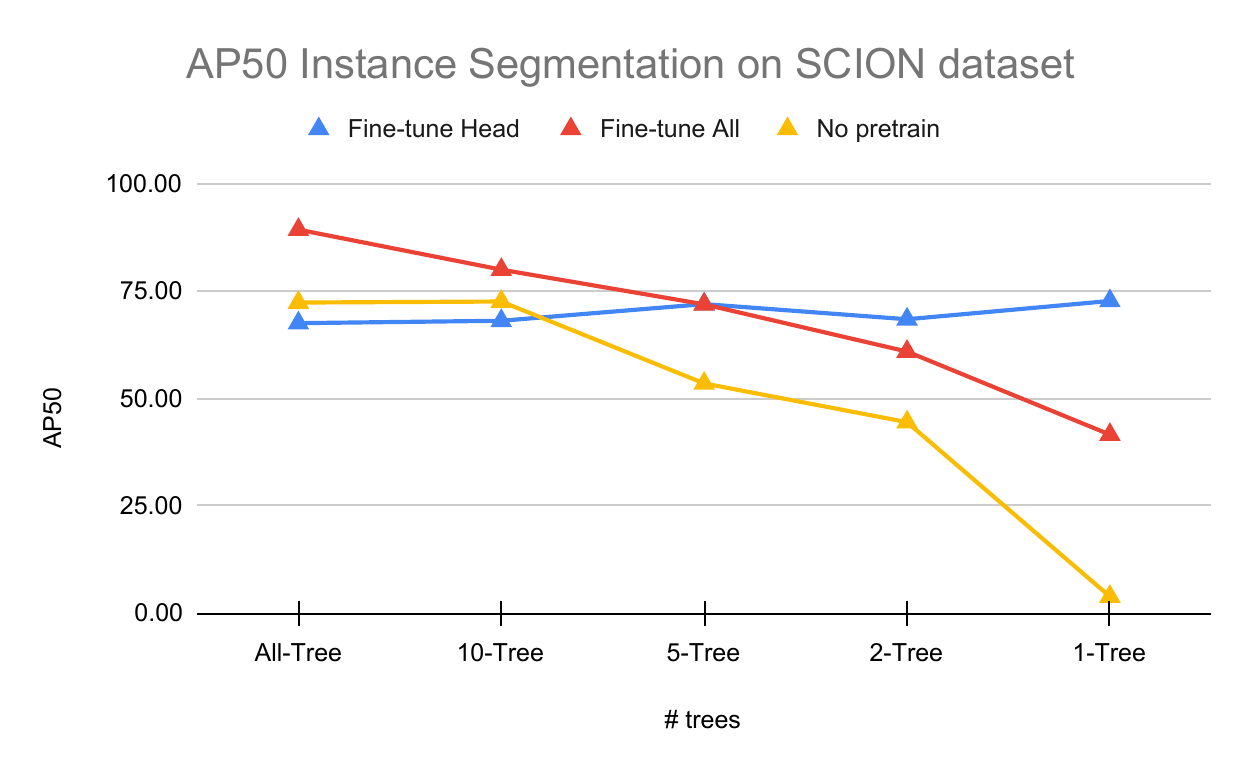}
         \caption{}
     \end{subfigure}
     \caption{Performance of instance segmentation under tree-level label reduction on the SCION dataset.}
     \label{fig:insseg_treelevel}
\end{figure*}

The trend in the entire experiments can be seen more clearly in Figure \ref{fig:insseg_uniform} and \ref{fig:insseg_treelevel}, where the performance of mAP and AP50 is visualized. The charts highlight how pretrained models maintain higher accuracy across all supervision levels, while the scratch-trained model degrades more sharply as labeled data become extremely sparse, e.g. 0.1\%.

\begin{figure*}[ht!]
     \centering
     \begin{subfigure}[b]{0.24\textwidth}
         \centering
         \includegraphics[width=\textwidth]{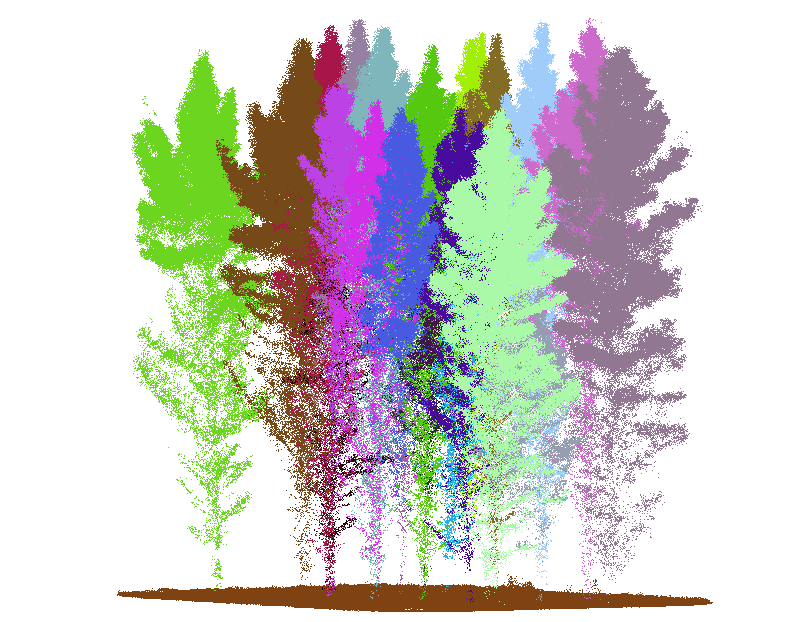}
         \caption{1\% fine-tuned all layers}
     \end{subfigure}
     \begin{subfigure}[b]{0.24\textwidth}
         \centering
         \includegraphics[width=\textwidth]{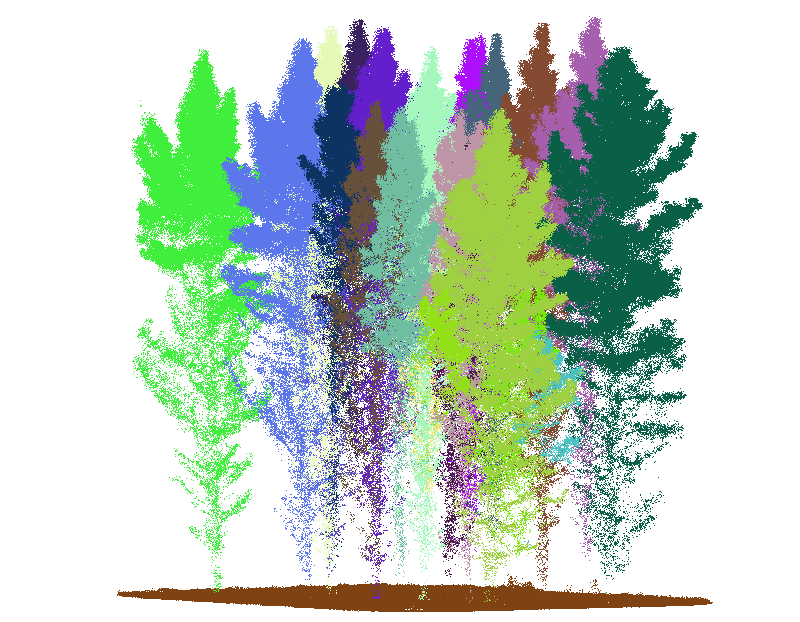}
         \caption{0.1\% fine-tuned all layers}
     \end{subfigure}
     \begin{subfigure}[b]{0.24\textwidth}
         \centering
         \includegraphics[width=\textwidth]{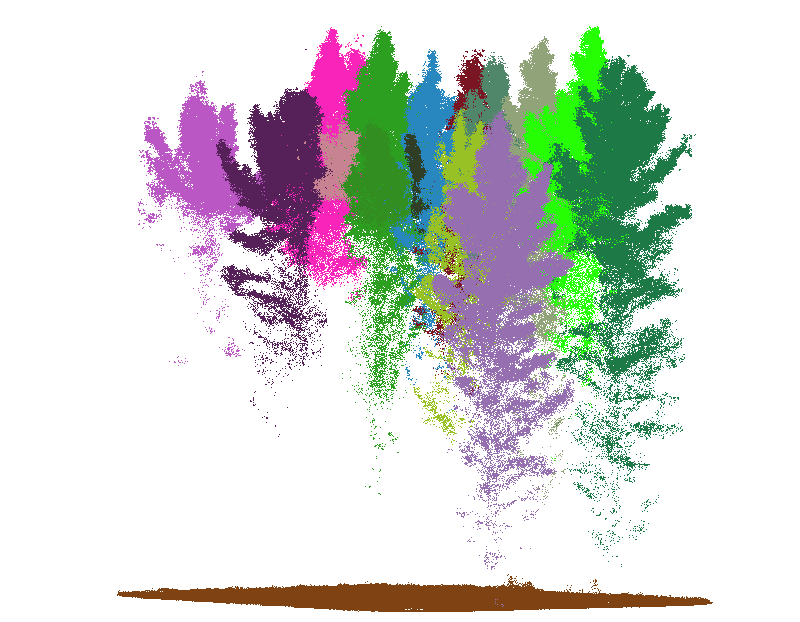}
         \caption{0.01\% fine-tuned all layers}
     \end{subfigure}
     \begin{subfigure}[b]{0.24\textwidth}
         \centering
         \includegraphics[width=\textwidth]{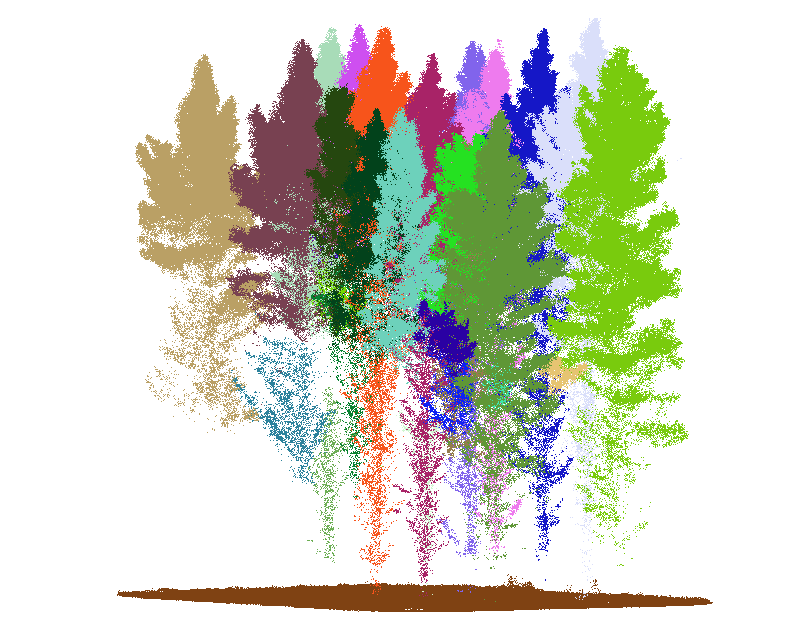}
         \caption{0.01\% fine-tuned head layers}
     \end{subfigure}
     \begin{subfigure}[b]{0.24\textwidth}
         \centering
         \includegraphics[width=\textwidth]{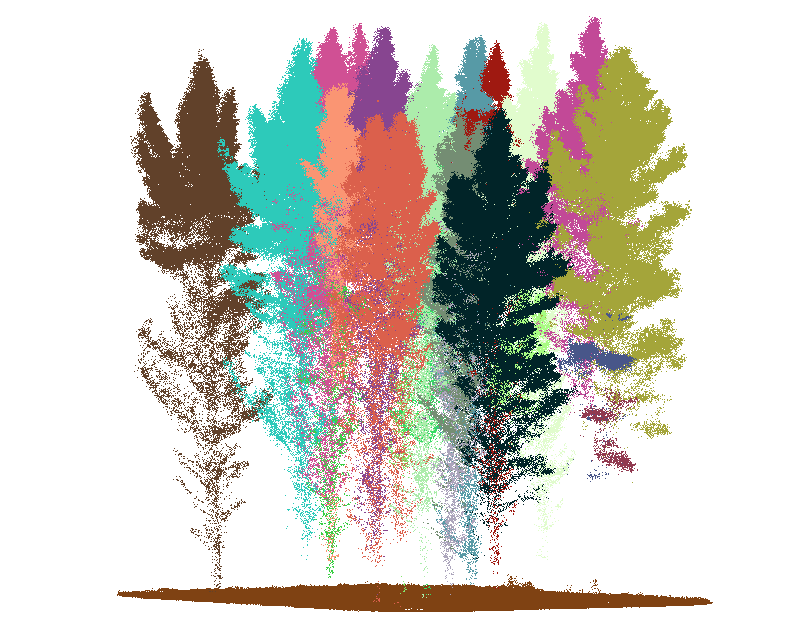}
         \caption{1\% from-scratch}
     \end{subfigure}
     \begin{subfigure}[b]{0.24\textwidth}
         \centering
         \includegraphics[width=\textwidth]{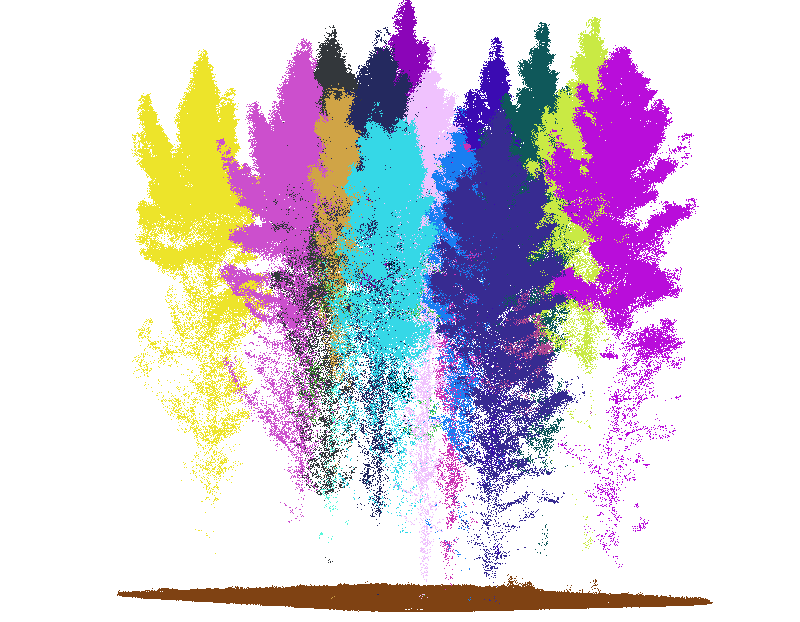}
         \caption{0.1\% from-scratch}
     \end{subfigure}
     \begin{subfigure}[b]{0.24\textwidth}
         \centering
         \includegraphics[width=\textwidth]{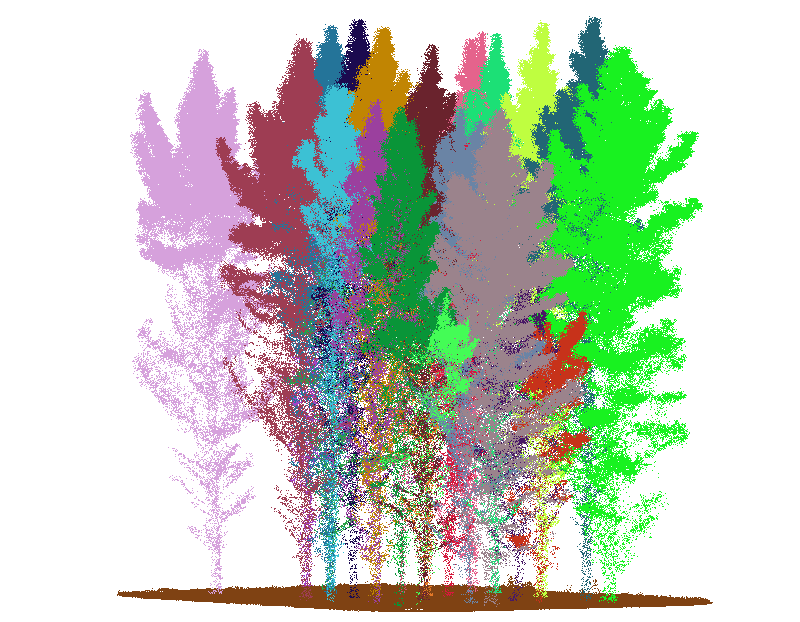}
         \caption{ground truth}
     \end{subfigure}
     \caption{Instance segmentation results under different label proportions.}
     \label{fig:insseg_preds_dataefficient}
\end{figure*}

Figure \ref{fig:insseg_preds_dataefficient} further shows qualitative comparisons of instance segmentation on the SCION region across different label proportions (1\%, 0.1\%, and 0.01\%) and training strategies. The top row shows results from models pretrained with self-supervised training and fine-tuned with domain adaptation. The 0.01\% setting includes both full and head-only fine-tuning. The bottom row shows from-scratch baselines and the ground truth. Results for 0.01\% from-scratch training are not shown, as the model fails to predict any instances from the point clouds.

At 1\% and 0.1\% label levels, models trained with self-supervised learning and domain adaptation produce reliable instance predictions, while from-scratch models often miss parts of instances, particularly in the lower sections of the trees. 

Under the extreme 0.01\% setting, full-fine-tuning of the pretrained model predicts some instances but leaves many points unlabeled. Head-only tuning captures a greater portion of the instances. In contrast, from-scratch training completely fails to detect any instances.

\subsection{Semantic segmentation results}

Table \ref{tab:semseg_100_01} shows the semantic segmentation performance on the FOR-instance dataset, comparing models initialized with self-supervised pretrained weights against those trained from scratch. We report results from two scenarios: first, a fully supervised setting where all available labels are used, and second, a data-efficient setting where only 0.1\% of the labeled data is available to simulate a few-shot scenario. 

\begin{table*}[ht!]
    \centering
    \small
    \setlength{\tabcolsep}{3pt} 
    \caption{Semantic segmentation performance (IoU and accuracy) using full supervision (100\%) and limited supervision (0.1\%) on all forest regions of FOR-instance dataset}
    
    % === Full Supervision ===
    %\subcaption*{\textbf{(a) Full Supervision (100\%)}}
    \resizebox{\textwidth}{!}{%
    \begin{tabular}{c|cc|cc|cc|cc}
        \hline
         \multirow{3}{*}{\textbf{Class}} & \multicolumn{4}{c|}{\textbf{Full supervision (100\%)}} &  \multicolumn{4}{c}{\textbf{Limited supervision (0.1\%)}}\\
         \cline{2-9}
       % \hline
       % \multirow{2}{*}{Class}
        & \multicolumn{2}{c|}{Pretrained} 
        & \multicolumn{2}{c|}{From Scratch}
        & \multicolumn{2}{c|}{Pretrained}
        & \multicolumn{2}{c}{From Scratch} \\
        \cline{2-9}
        & IoU (\%) & Accuracy (\%) 
        & IoU (\%) & Accuracy (\%)
        & IoU (\%) & Accuracy (\%)
        & IoU (\%) & Accuracy (\%) \\
        \hline
        stem    & 64.85 & 74.63 & 63.95 & 74.33 & 61.41 & 71.69 & 61.84 & 72.67 \\
        crown   & 91.80 & 96.22 & 90.30 & 94.42 & 90.78 & 95.02 & 89.33 & 93.76 \\
        branch  & 59.61 & 75.84 & 55.65 & 78.06 & 55.64 & 75.33 & 52.98 & 76.22 \\
        terrain & 96.75 & 97.49 & 95.94 & 96.63 & 97.38 & 98.31 & 95.22 & 96.05 \\
        \textbf{mean} & \textbf{78.25} & \textbf{86.04} & \textbf{76.46} & \textbf{85.86} & \textbf{76.30} & \textbf{85.14} & \textbf{74.84} & \textbf{84.77} \\
        \hline
    \end{tabular}}
    \label{tab:semseg_100_01}
\end{table*}

Pretraining consistently improves performance in both scenarios, but the gains are most noticeable under limited supervision. Notably, with just 0.1\% of labeled data, the pretrained model achieves mean IoU of 76.30\%, which is comparable to the 76.46\%, achieved by a model trained from scratch with full supervision.

\begin{figure}[ht!]
    \centering
    \includegraphics[width=0.9\linewidth]{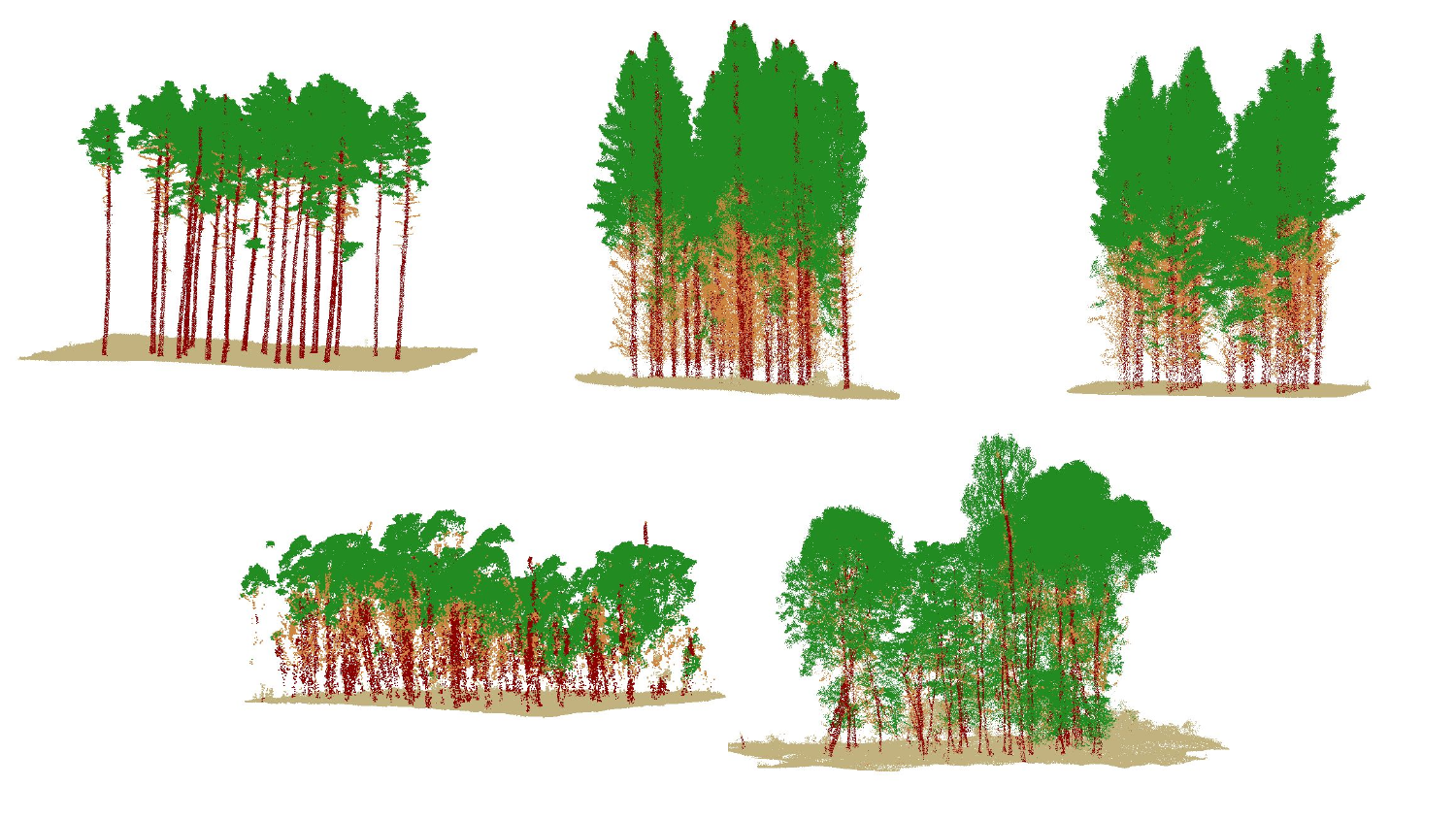}
    \caption{Visualization of semantic segmentation results on the different forest regions in the FOR-instance dataset.}
    \label{fig:all_semseg_results}
\end{figure}

We can further observe qualitatively in Figure \ref{fig:all_semseg_results}, which presents the visualization of semantic segmentation predictions across different forest regions in the FOR-instance dataset.

\subsection{Tree classification results}
\label{sec:tree_classification_results}
Table \ref{tab:classification_pretrain} presents classification results on two broad tree categories, conifer and broadleaf, used as a pretrained task. The pretrained model is then fine-tuned for species-level classification, as shown in Table \ref{tab:classification_species}, where results are also compared to from-scratch training. 

\begin{table}[ht]
    \centering
    \caption{Tree classification validation performance.}
    \begin{tabular}{lcc}
        \hline
        \textbf{Metric / Class} & \textbf{Jaccard} & \textbf{Accuracy} \\
        \hline
        \textbf{Mean (mJaccard / mAcc)} & 0.7464 & 0.8569 \\
        \hline
        Broadleaf & 0.8207 & 0.8915 \\
        Conifer   & 0.6721 & 0.8223 \\
        \hline
    \end{tabular}
\label{tab:classification_pretrain}
\end{table}

\begin{table*}[ht!]
    \centering
    \caption{Tree classification performance (Jaccard index and accuracy) comparison on species-level  using few-shot learning.}
    \resizebox{\textwidth}{!}{%
    \begin{tabular}{c|cc|cc}
        \hline
        \multirow{2}{*}{\textbf{Species}}
        & \multicolumn{2}{c|}{Pretrained} 
        & \multicolumn{2}{c}{From Scratch} \\
        \cline{2-5}
        & Jaccard (\%) & Accuracy (\%) 
        & Jaccard (\%) & Accuracy (\%) \\
        \hline
        \textit{Pinus sylvestris} & 86.67 & 95.12 & 80.43 & 90.24 \\
        \textit{Quercus robur}    & 85.37 & 89.74 & 79.07 & 87.18 \\
        \textbf{mean}    & \textbf{86.02} & \textbf{92.43} & \textbf{79.75} & \textbf{88.71} \\
        \hline
        \textit{Pinus sylvestris} & 58.40 & 66.25 & 53.57 & 77.40 \\
        \textit{Picea abies}    & 64.72 & 86.56 & 45.28 & 55.52 \\
        \textbf{mean}    & \textbf{61.56} & \textbf{76.41} & \textbf{49.43} & \textbf{66.46} \\
        \hline
        \textit{Quercus robur} & 38.06 & 54.55 & 23.35 & 62.57 \\
        \textit{Acer campestre}    & 84.13 & 91.57 & 62.75 & 67.33 \\
        \textbf{mean}    & \textbf{61.09} & \textbf{73.06} & \textbf{43.05} & \textbf{64.95} \\
        \hline
    \end{tabular}}
    \label{tab:classification_species}
\end{table*}
    
In the few-shot setting with limited labeled data, pretraining provides a substantial boost. For inter-group classification (\textit{Pinus sylvestris} vs. \textit{Quercus robur}), the mean Jaccard index increases from 79.95\% to 86.02\%. For intra-group classification, the mean Jaccard index rises from 49.43\% to 61.56\% for \textit{Pinus sylvestris} vs. \textit{Picea abies}, and from 43.05\% to 61.09\% for \textit{Quercus robur} vs. \textit{Acer campestre}. Note that intra-group classifications resulted in lower accuracies, both with from-scratch training and with pretrained weights, reflecting the greater difficulty of distinguishing between species within the same taxonomic group. 

Overall, this demonstrates that initializing the model with a coarse-level classification task (broadleaf vs. conifer) leads to more effective feature learning, significantly improving downstream species classification performance compared to training from scratch.

\subsection{Evaluation with state-of-the-art}
We evaluated the performance of our method against ForAINet (\cite{XIANG2024114078}), SegmentAnyTree (SAT)(\cite{WIELGOSZ2024114367}), and Treeiso (\cite{rs14236116}) across multiple test sites using standard instance segmentation metrics, including F1-score, precision, and recall, as well as commission and omission errors, along with the TP, the FP, and the FN.

Table \ref{tab:insseg_all_comp} indicates that our model achieves the highest precision among the three deep-learning-based methods, reflecting a lower rate of false positive detections. This characteristic is important for applications where minimizing incorrect identifications is critical. ForAINet demonstrates a slightly higher recall and F1-score, suggesting a greater ability to detect true instances. SAT shows balanced performance with moderate precision and recall, accompanied by a relatively higher commission rate. These differences illustrate a trade-off between precision and recall across the methods, with each approach emphasizing different aspects of detection performance. 

\begin{table*}[ht!]
    \centering
    \small
    \setlength{\tabcolsep}{3pt} 
    \caption{Instance segmentation performance across different forest types against state-of-the-art approaches}
    %\rowcolors{3}{}{lightgray}
    \resizebox{\textwidth}{!}{%
    \begin{tabular}{c|c|c|c|c|c|c|c|c|c}
        \hline
        \multirow{2}{*}{\textbf{Forest types}} & \multirow{2}{*}{\textbf{Methods}} & \textbf{F1} & \textbf{Precision} & \textbf{Recall} & \textbf{Commiss.} & \textbf{Omiss.} & \textbf{TP} & \textbf{FP}& \textbf{FN} \\
         &  & (\%) $\uparrow$ & (\%) $\uparrow$ & (\%) $\uparrow$ & \textbf{Err.} (\%) $\downarrow$ & \textbf{Err.} (\%) $\downarrow$ & $\uparrow$ & $\downarrow$ & $\downarrow$ \\
        \hline
        \multirow{4}{*}{CULS} & Ours & 100.0 & 100.0 & 100.0 & 0.0 & 0.0 & 20 & 0 & 0 \\
         & ForAINet & 93.0 & 87.0 & 100.0 & 13.0 & 0.0 & 20 & 3 & 0 \\
         & SAT & 99.0 & 100.0 & 100.0 & 0.0 & 0.0 & -- & -- & -- \\
         & Treeiso & 90.0 & 90.0 & 90.0 & 10.0 & 10.0 & 18 & 2 & 2 \\
        \hline
        \multirow{4}{*}{NIBIO} & Ours & 90.9 & 98.5 & 84.5 & 1.4 & 15.5 & 136 & 2 & 25 \\
         & ForAINet & 91.9 & 96.6 & 87.6 & 3.4 & 12.4 & 141 & 5 & 20 \\
         & SAT & 88.0 & 91.0 & 88.0 & 9.0 & 12.0 & -- & -- & -- \\
         & Treeiso & 79.4 & 92.6 & 69.6 & 7.4 & 30.4 & 112 & 9 & 49 \\
        \hline
        \multirow{4}{*}{RMIT} & Ours & 70.9 & 84.8 & 60.9 & 15.2 & 39.0 & 39 & 7 & 25 \\
         & ForAINet & 69.5 & 75.9 & 64.1 & 24.1 & 35.9 & 41 & 13 & 23 \\
         & SAT & 83.0 & 69.0 & 83.0 & 17.0 & 31.0 & -- & -- & -- \\
         & Treeiso & 36.2 & 46.3 & 29.6 & 53.7 & 70.3 & 19 & 22 & 45\\
        \hline
        \multirow{4}{*}{SCION} & Ours & 91.1 & 100.00 & 83.7 & 0.0 & 16.3 & 36 & 0 & 7 \\
         & ForAINet & 91.6 & 95.0 & 98.4 & 5.0 & 11.6 & 38 & 2 & 5 \\
         & SAT & 91.0 & 93.0 & 92.0 & 7.0 & 8.0 & -- & -- & -- \\
         & Treeiso & 77.1 & 80.0 & 74.4 & 20.0 & 25.6 & 32 & 8 & 11 \\
        \hline
        \multirow{4}{*}{TU-WIEN} & Ours & 54.5 & 75.0 & 42.8 & 25.0 & 57.1 & 15 & 5 & 20 \\
         & ForAINet & 69.4 & 67.6 & 71.4 & 32.4 & 28.6 & 25 & 12 & 10 \\
         & SAT & 57.0 & 55.0 & 46.0 & 45.0 & 54.0 & -- & -- & -- \\
         & Treeiso & 29.7 & 28.2 & 31.4 & 71.8 & 68.6 & 11 & 28 & 24 \\
        \hline
    \end{tabular}}
    \label{tab:insseg_all_comp}
\end{table*}

Compared to a non-deep learning based method, Treeiso, our model achieves superior performance. Treeiso struggles to achieve optimum performance in dense forests and tends to produce undersegmented predictions, indicated by high false negative numbers, which leads to poor recall values. Treeiso does not require labeled data, which is a main advantage, only requiring the user to tune the parameters, however we found that parameter tuning is not really straightforward.

\section{Discussion and Analysis}
\label{sec:discussion}

\subsection{Performance across diverse forests}
Our instance segmentation results demonstrate the model's strong capability to detect tree instances across diverse forest environments. In structured forest areas like CULS, NIBIO, and SCION, performance is high, with many test scenes yielding zero false positives and precision reaching 100\%. In more heterogeneous and challenging domains such as RMIT and TU WIEN, precision remains acceptable ($>$ 75\%), indicating the model's conservative nature, rarely over-segmenting tree instances. However, these regions also exhibit significantly lower recall due to misdetections. 

We attribute this to structural variability and lower point density, which jointly affect the offset prediction and clustering stages in the PointGroup framework. Structural variability influences instance segmentation by reducing the geometric separability between neighboring trees. In more complex forests, irregular crown shapes introduce ambiguous local geometry, making it difficult for the model to learn consistent point-wise offsets. As a result, points from adjacent trees may be shifted toward similar regions in the offset space.

During the BFS clustering stage, which relies on a fixed distance threshold, these ambiguities lead to incorrect connectivity between neighboring trees. In particular, the fixed threshold may connect points belonging to different trees instead of separating them into distinct instances. This results in multiple trees being merged into a single predicted instance (under segmentation). This behavior is consistent with our observations, where the number of predicted instances is lower than the ground truth, leading to a high number of FN.

Point density further amplifies this issue. RMIT and TUWIEN have the lowest point densities, with approximately 498 and 1,717 points/$m^2$, respectively, whereas sites such as NIBIO reach around 9,500 points/$m^2$. This sparse sampling weakens local geometric cues and reduces the effective separation between neighboring trees after offset shifting, increasing the likelihood of incorrect connections during clustering. As a result, structurally complex forests combined with low point density leads to increased instance merging and higher FN rates. These findings are consistent with observations reported by \cite{XIANG2024114078}, who also showed that forests with complex crowns increase ambiguity and segmentation difficulty.

This highlights a trade-off inherent to offset based clustering approaches, where inaccuracies in offset prediction combined with a fixed BFS radius can lead to missed instances, suggesting the need for more adaptive grouping strategies to improve recall. 

\subsection{Instance segmentation under sparse supervision}

Instance segmentation under extremely sparse supervision remains challenging, as direct fine-tuning often fails to learn meaningful point shifts for clustering. Domain adaptation bridges the gap by exposing the model to labeled data from a related source domain before fine-tuning, improving few-shot performance. We also adapt a two-stage fine-tuning scheme to gradually specialize the pretrained features.

We evaluate this approach with training from scratch across different labeled proportions using SCION as an unseen test site. When more than 0.1\% labels are available, full fine-tuning achieves the best performance by allowing domain-specific adaptation. However, at extremely low labeled density (0.01\%), full fine-tuning overfits, while head-only tuning is more robust by preserving the rich features of the pretrained backbone.

In the head-only fine-tuning setting, it consistently outperforms training from scratch across label levels, confirming that pretrained backbone and domain adaptation provide strong, transferable representations. In contrast, random initialization leads to unstable convergence and poor generalization.

\begin{figure*}[ht!]
     \centering
     \begin{subfigure}[b]{0.45\textwidth}
         \centering
         \includegraphics[width=\textwidth]{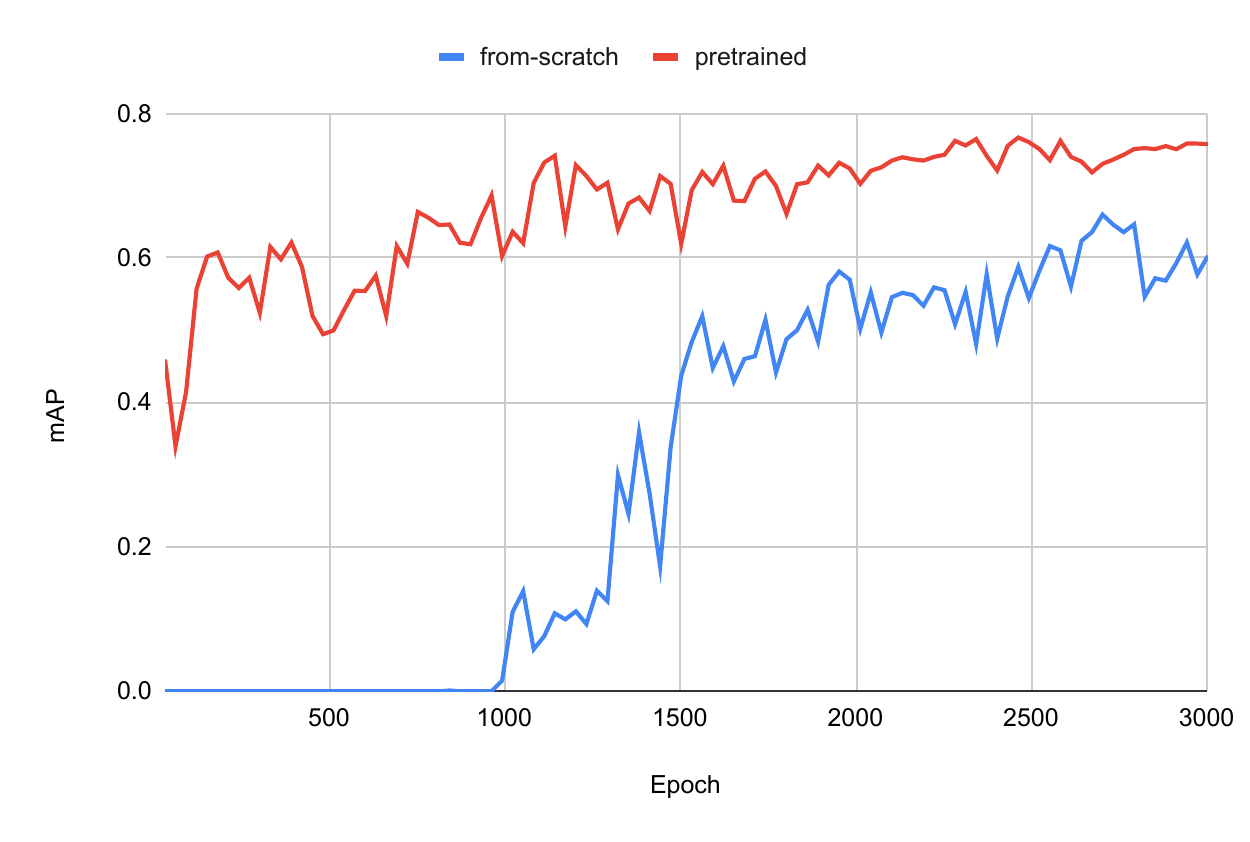}
         \caption{mAP (full supervision)}
     \end{subfigure}
     \begin{subfigure}[b]{0.45\textwidth}
         \centering
         \includegraphics[width=\textwidth]{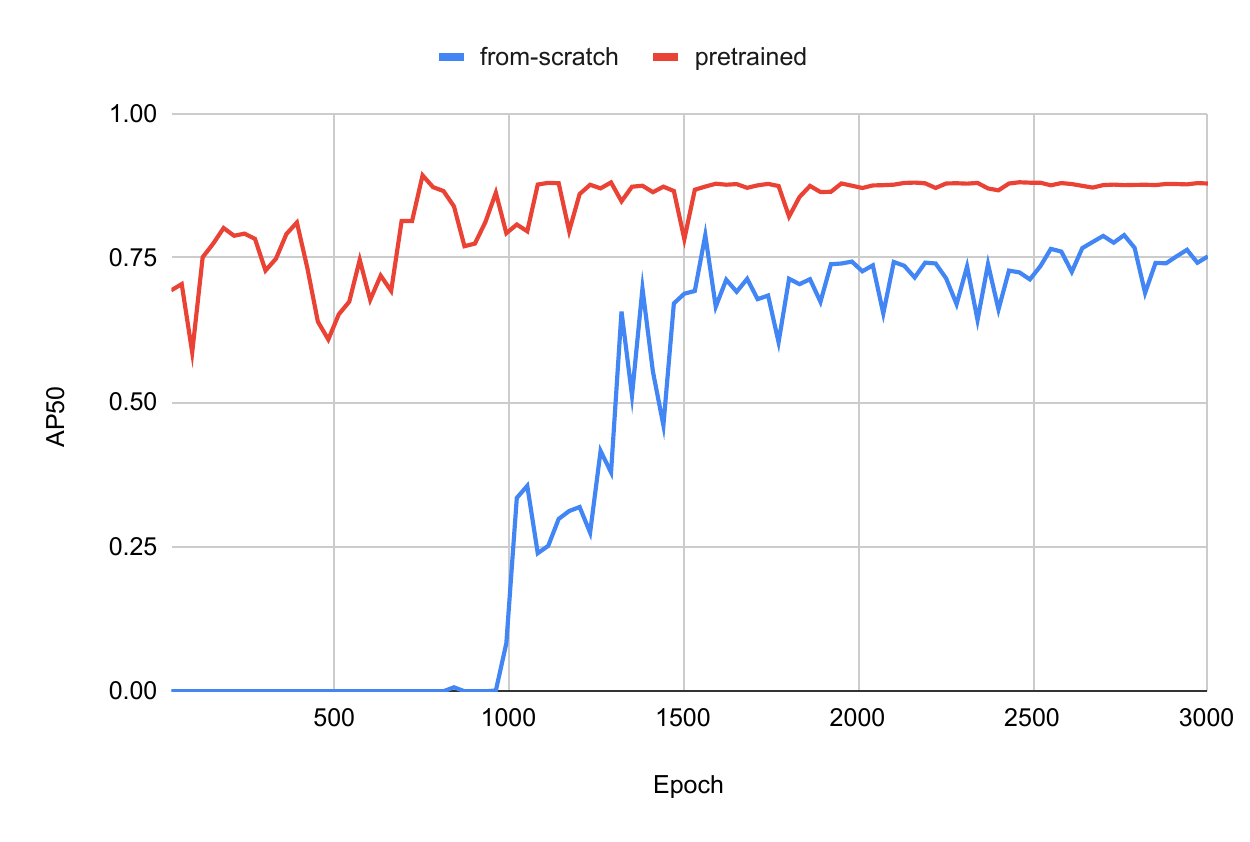}
         \caption{AP50 (full supervision)}
     \end{subfigure}
     \begin{subfigure}[b]{0.45\textwidth}
         \centering
         \includegraphics[width=\textwidth]{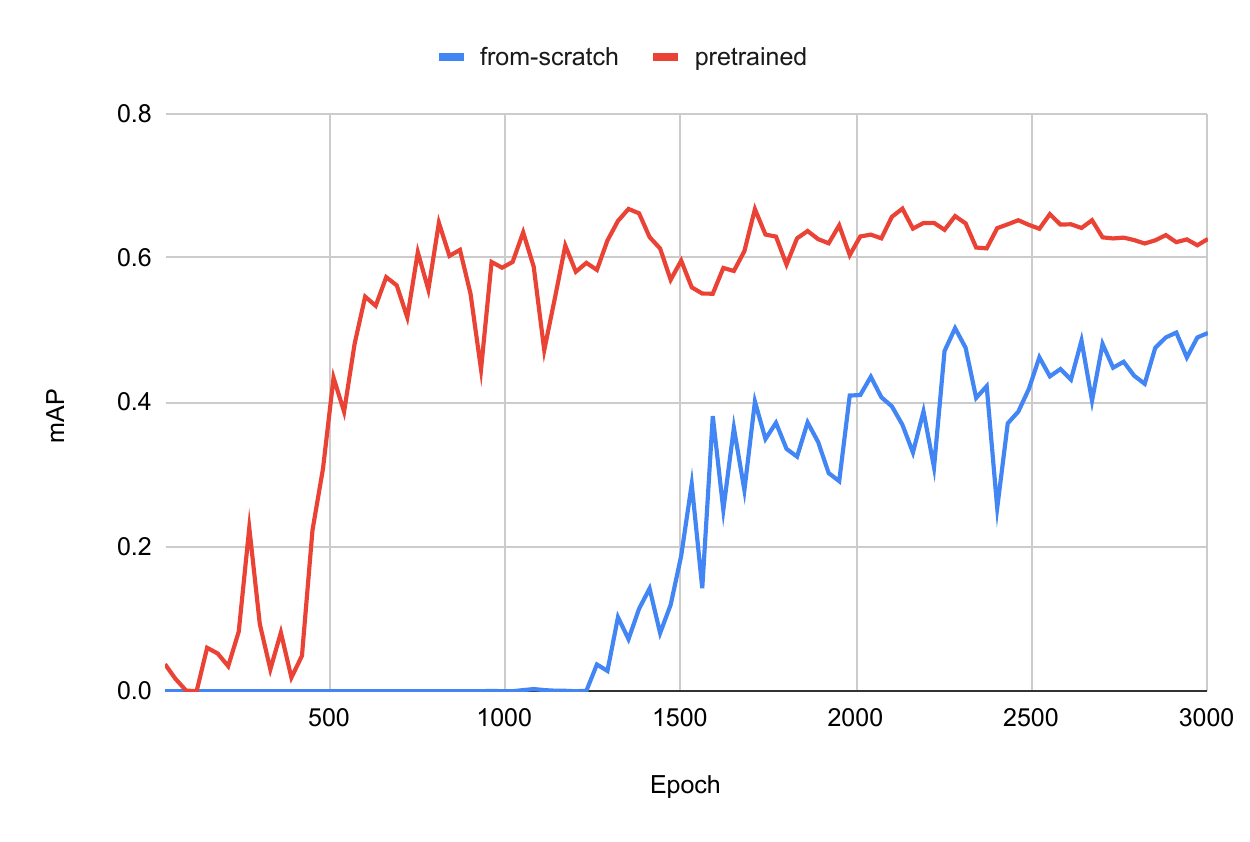}
         \caption{mAP (0.1\% labels)}
     \end{subfigure}
     \begin{subfigure}[b]{0.45\textwidth}
         \centering
         \includegraphics[width=\textwidth]{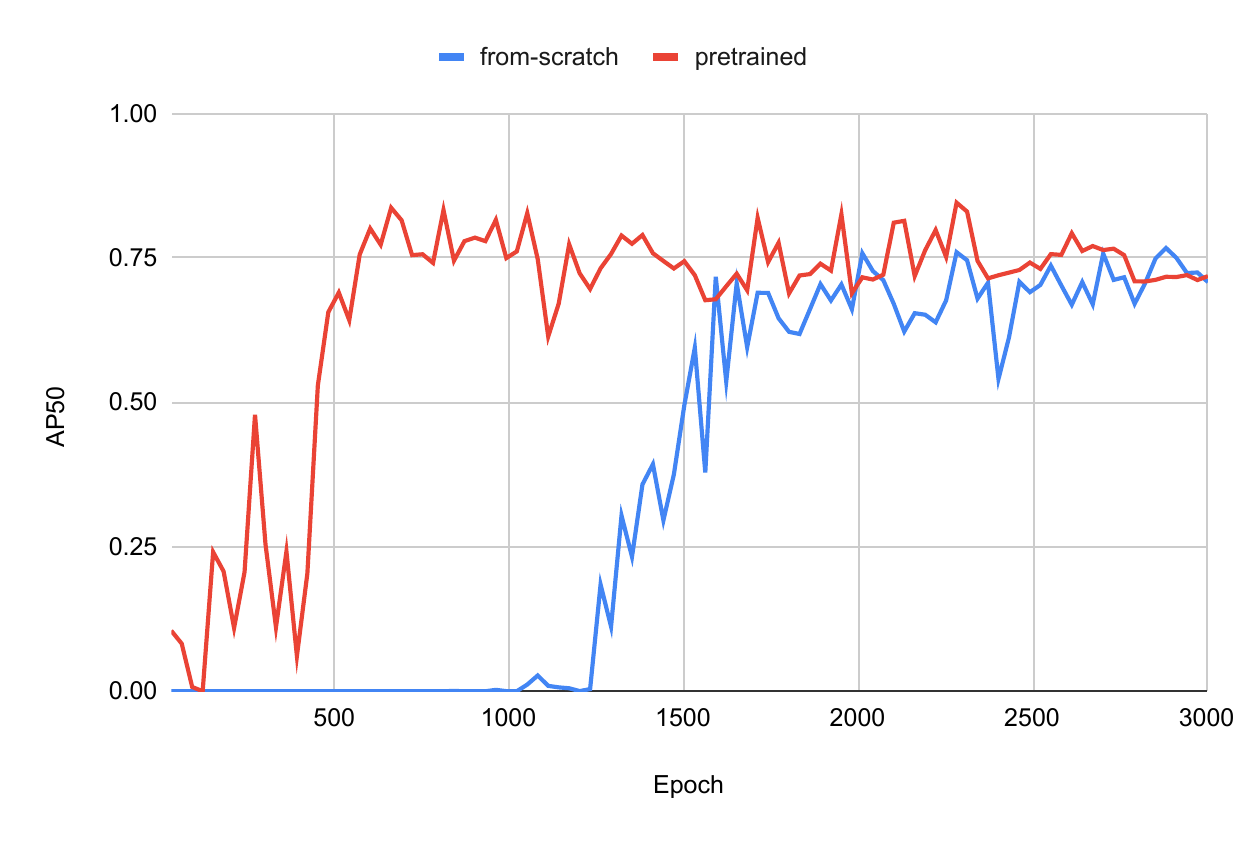}
         \caption{AP50 (0.1\% labels)}
     \end{subfigure}
     \caption{Validation curves of mAP and AP50 under full supervision (a and b) and limited supervision (c and d) during training. Models initialized with self-supervised and domain adaptation pretrained weights (\textcolor{red}{---}) converge faster and achieve higher accuracy, while models trained from scratch (\textcolor{blue}{---}), struggle to optimize in the nearly stages of training and show slower convergence.}
     \label{fig:insseg_curve}
\end{figure*}

Overall, these results support a progressive fine-tuning strategy: start with head-only updates under extremely label scarcity and switch to full fine-tuning when label availability increases. As shown in Figure \ref{fig:insseg_curve}, pretrained models converge faster and achieve higher mAP and AP50, demonstrating the advantage of strong initialization.

\subsection{Semantic segmentation under sparse supervision}
Unlike instance segmentation, semantic segmentation focuses on per-point classification, where even a few labeled points can provide sufficient supervision when supported by self-supervised pretraining. Domain adaptation is less critical, as the self-supervised encoder already captures transferable geometric and contextual features that generalize well across domains. Since the task involves labeling individual points rather than grouping them into objects, the general-purpose features learned during pretraining are often sufficient.

\begin{figure*}[ht!]
     \centering
     \begin{subfigure}[b]{0.45\textwidth}
         \centering
         \includegraphics[width=\textwidth]{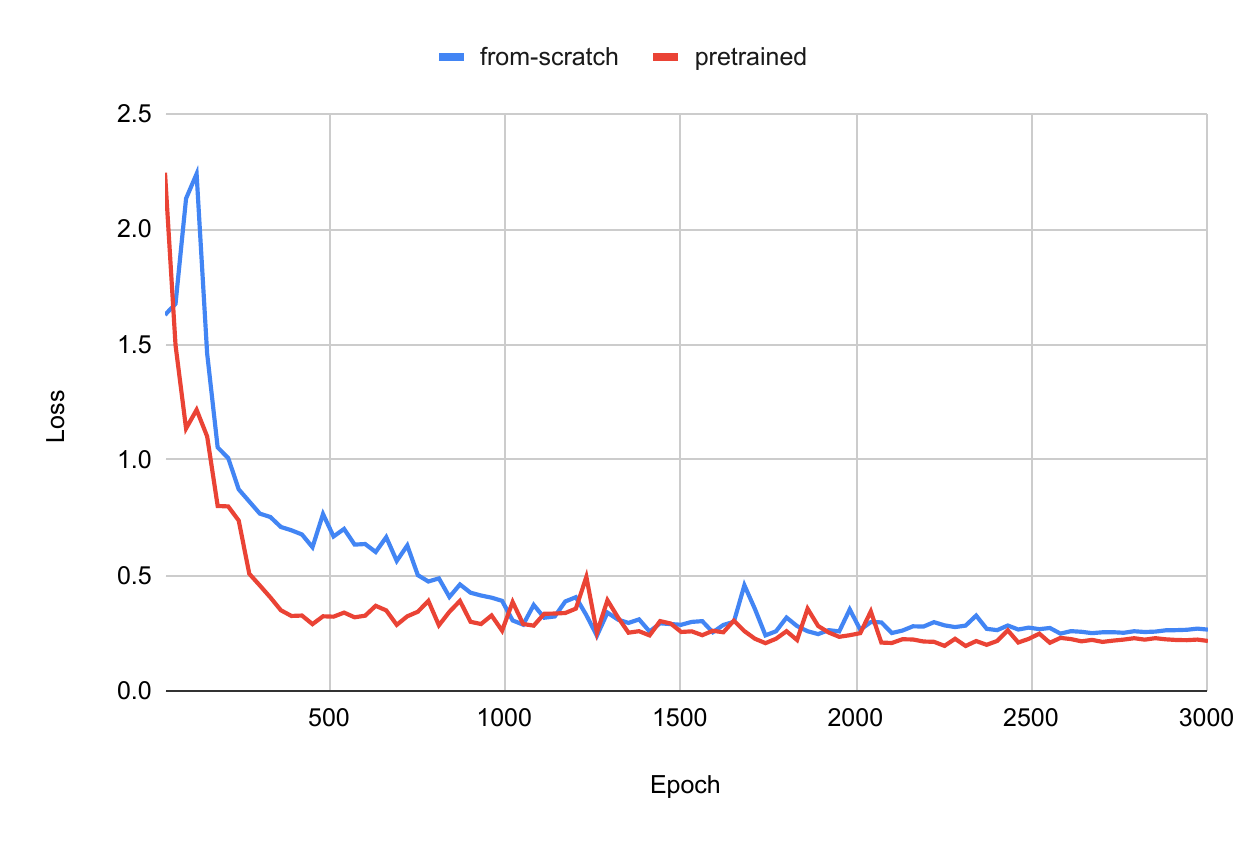}
         \caption{Validation loss (full supervision)}
     \end{subfigure}
     \begin{subfigure}[b]{0.45\textwidth}
         \centering
         \includegraphics[width=\textwidth]{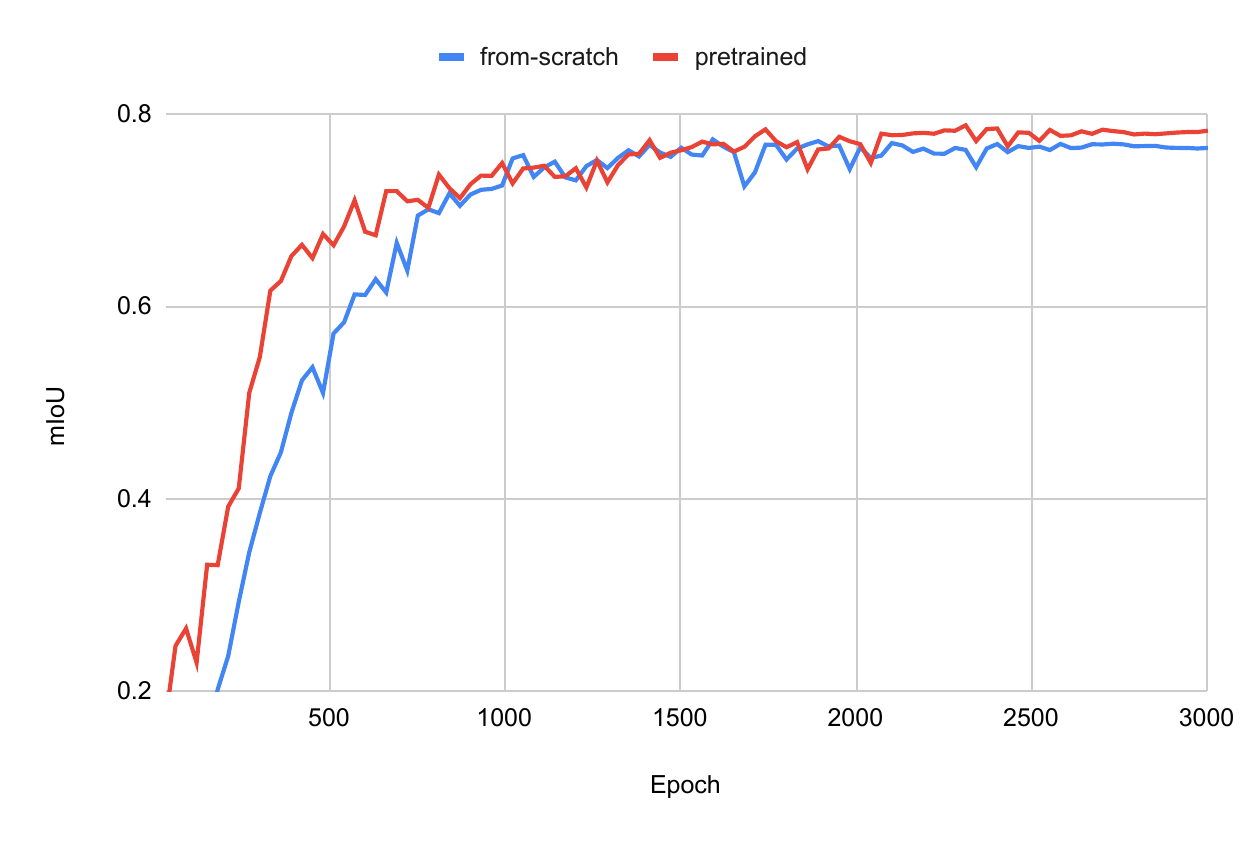}
         \caption{mIoU (full supervision)}
     \end{subfigure}
     \begin{subfigure}[b]{0.45\textwidth}
         \centering
         \includegraphics[width=\textwidth]{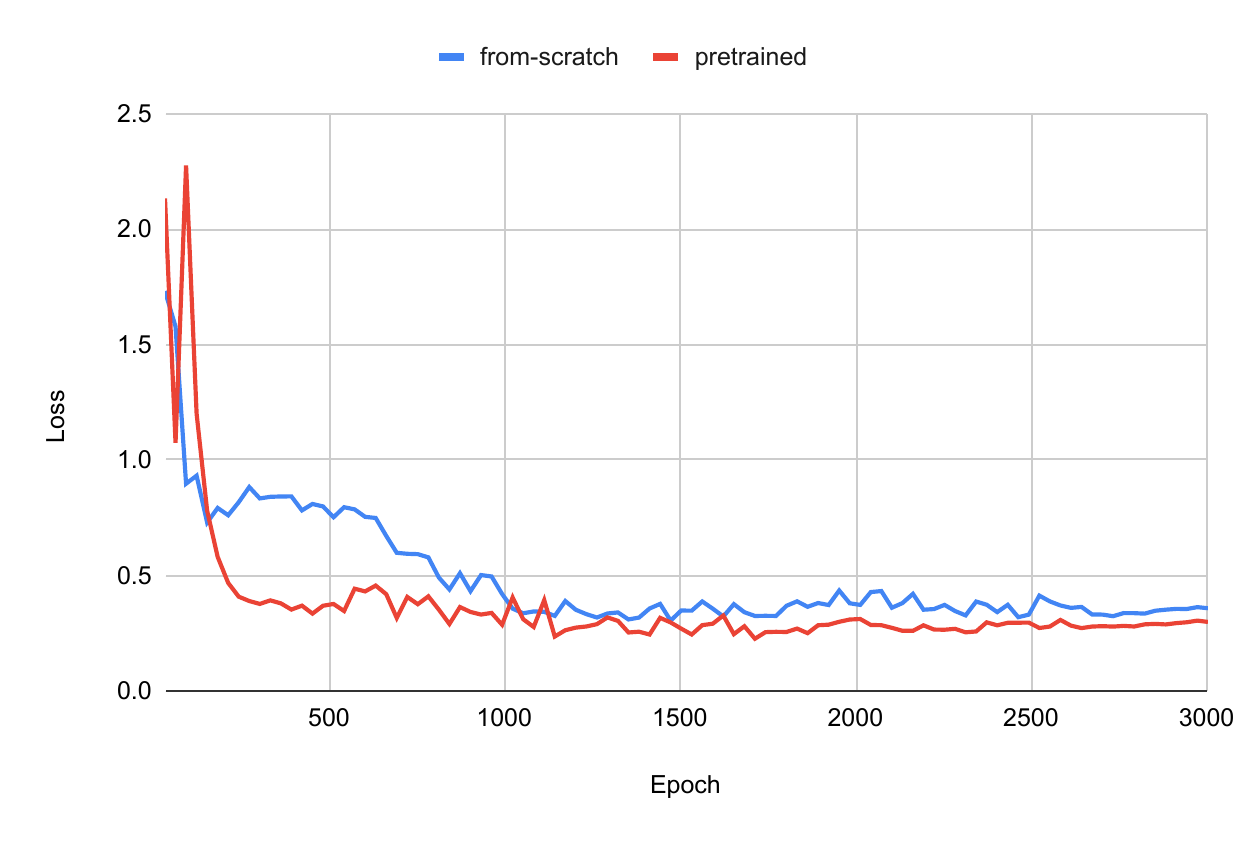}
         \caption{Validation loss (0.1\% labels)}
     \end{subfigure}
     \begin{subfigure}[b]{0.45\textwidth}
         \centering
         \includegraphics[width=\textwidth]{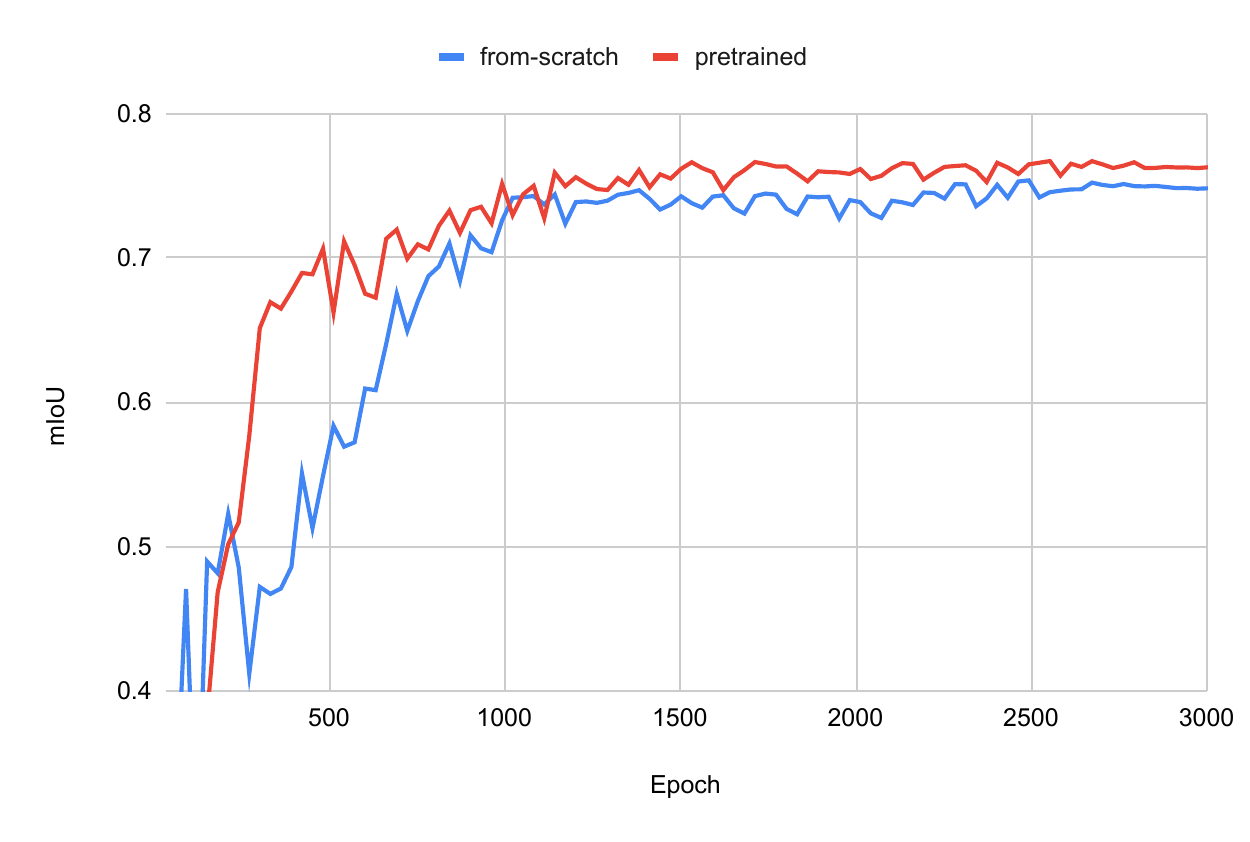}
         \caption{mIoU (0.1\% labels)}
     \end{subfigure}
     \caption{Validation curves of loss and mIoU (a and b) during training showing that models initialized with self-supervised pretrained weights (\textcolor{red}{---}) converge faster and achieve better early performance compared to training from scratch (\textcolor{blue}{---}), especially under limited supervision (c and d).}
     \label{fig:semseg_curve}
\end{figure*}

Figure \ref{fig:semseg_curve} shows the validation mIoU and loss for models initialized with pretrained versus random weights. The pretrained model converges faster and achieves higher accuracy, particularly under limited supervision, indicating that self-supervised features capture structural and contextual information that remains effective even with minimal labels. 

Overall, semantic segmentation consistently benefits from self-supervised pretraining, while instance segmentation demands additional task-specific adaptation due to the complexity of learning object-level grouping.

\subsection{Impact of label reduction strategies}
We evaluate two strategies to reduce annotation effort for the instance segmentation task: (1) uniform label reduction across many trees, and (2) tree-level label reduction where few trees are fully labeled. In a situation where both have a similar total number of labeled points, the sparse-and-diverse approach (uniform label reduction) consistently outperforms the dense-and-few strategy (tree-level label reduction). 

This is likely because labeling a wide variety of trees exposes the model to a greater diversity of tree structures. This leads to a better generalization. On the other hand, densely labeling a small number of trees often results in redundant information, as neighboring points on the same tree tend to have similar representations and contribute less information to the model. 

%We also observe that the model's performance remains stable under uniform label reduction, even when only 1\% of labeled data is available. However, performance immediately drops when the number of unique labeled tree instances is reduced. This suggests that \textcolor{blue}{instance-level task learning primarily relies on instance count, not label density,} as sparse annotations are sufficient for the model to learn reliable instance-level features and offset vectors for instance grouping via the BFS clustering algorithm. While extremely low proportions of labeled data are indeed affecting performance, self-supervised pretraining and domain adaptation help the model maintain a reasonable ability to predict tree instances. 

These findings highlight that the diversity of labeled examples is more valuable than the density of labels per example. From a practical standpoint, this has important implications. When annotation resources are limited, it is more effective to sparsely label many trees rather than exhaustively label a few. In other words, human annotators don't need to label every point fully, but labeling enough distinct trees is crucial to achieve satisfactory performance.

However, it is likely that the sparse labels must still adequately represent the overall shape of the tree, as the uniform random sampling maintains structural integrity even at 1\% level. This contrasts with real-world annotation, where human annotators typically prefer to label only a few point groups in different parts of the tree for faster labeling, rather than uniformly sparse points. Therefore, whether this would lead to similar results remains unclear in our experiment.  

%However, at the same time it is likely that the sparse labels still have to adequately represent the shape of the tree as was the case in our setting. It is important to mention this since sparse labeling could be indeed notably faster if only a few points in different parts of the tree would have to be labeled, however, whether this would lead to similar results remains unclear in our experiment. 

\subsection{Tree classification with few-shot learning}
As summarized in Section \ref{sec:tree_classification_results}, pretraining on broad tree categories before fine-tuning improves species classification performance. This approach benefits generalization to new species, particularly when only a small number of samples per species is available. This hierarchical strategy reduces the need for extensive labeled data, making it highly practical for real-world applications.

The improvement likely arises from the structural differences between broadleaf and coniferous trees. Conifers have narrow,  elongated crowns with dense needle-like foliage, and broadleaf trees show wider canopies, more complex branching structures. Pretraining on these broad groups helps the model to learn discriminative geometric and contextual features that generalize well to unseen species during fine-tuning. To avoid information leakage, target species used in fine-tuning are excluded from pretraining. However, in practical deployment, including more species during pretraining would enhance generalization.

\begin{figure*}[ht!]
     \centering
     \begin{subfigure}[b]{0.45\textwidth}
         \centering
         \includegraphics[width=\textwidth]{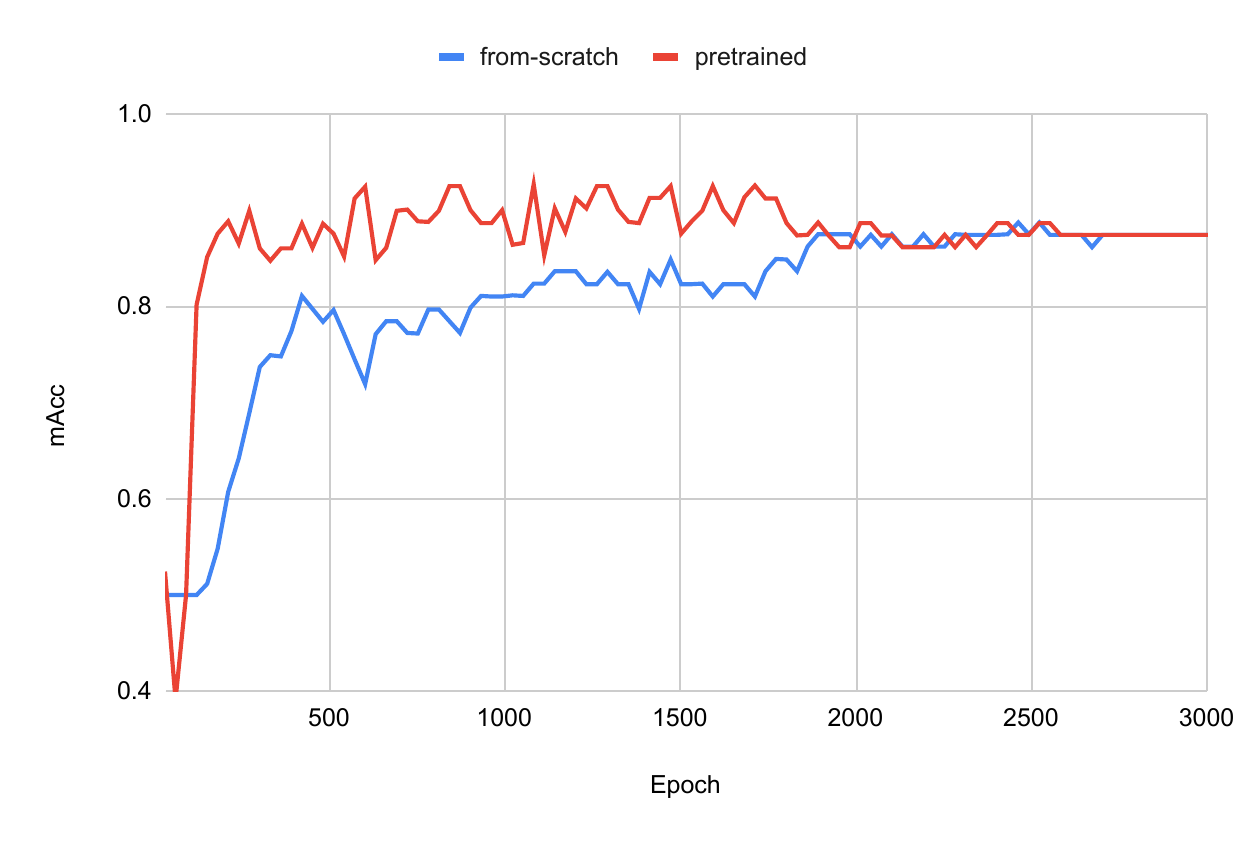}
         \caption{Mean accuracy}
     \end{subfigure}
     \begin{subfigure}[b]{0.45\textwidth}
         \centering
         \includegraphics[width=\textwidth]{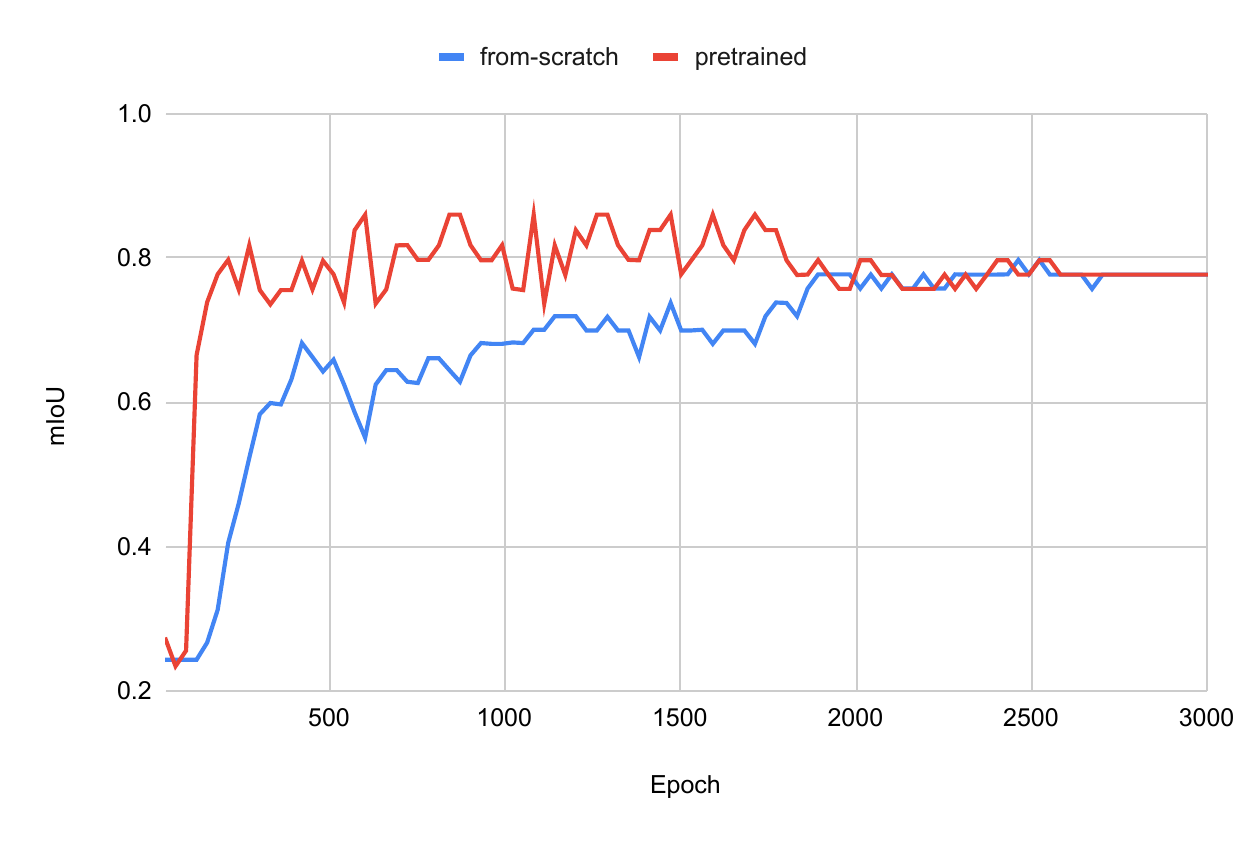}
         \caption{Mean IoU}
     \end{subfigure}
     \caption{Validation curves of mean accuracy (a) and mean IoU for few-shot learning of tree species classification. Models initialized with pretrained weights (\textcolor{red}{---}) converge faster and achieve highest accuracy during mid-training, while models trained from scratch (\textcolor{blue}{---}), converges much later.}
     \label{fig:classification_curve}
\end{figure*}

Figure \ref{fig:classification_curve} shows that pretrained models converge faster and reach peak performance in roughly half the training time. Models trained from scratch converge slower and still fall short in accuracy. Interestingly, prolonged training of pretrained models under few-shot learning can lead to slight overfitting, suggesting that early stopping or fewer epochs are preferable in few-shot settings.

\subsection{Energy consumption and environmental impact}

In addition to accuracy measurements, we use a CodeCarbon (\cite{benoit_courty_2024_11171501}) to monitor energy consumption and estimated carbon emissions during training on a single node equipped with four NVIDIA A100 GPUs. All experiments were conducted on the SCION dataset. Table \ref{tab:energy_comparison} summarizes measurements for models trained from scratch versus models fine-tuned from pretrained backbone, targeting similar instance segmentation accuracy.

\begin{table}[h!]
\centering
\caption{Energy consumption and estimated carbon emmissions for training from scratch versus fine-tuning with pretrained weights (achieving similar accuracy).}
\begin{tabular}{lcc}
    \hline
    \multirow{2}{*}{\textbf{Metric}} & \textbf{From} & \textbf{Pretrained} \\
     & \textbf{scratch} & \textbf{fine-tuning} \\
    \hline
    Duration (minutes) & 44.51 & 35.56 \\
    Energy consumed (kWh) & 0.5813 & 0.4605 \\
    Estimated emissions (kg CO$_2$) & 0.2215 & 0.1754 \\
    Average GPU Power (W) & 630.25 & 487.51 \\
    Average CPU Power (W) & 225.0 & 225.0 \\
    Average RAM Power (W) & 70.0 & 70.0 \\
\hline
\end{tabular}
\label{tab:energy_comparison}
\end{table}

Models initialized with pretrained weights converge faster to reach comparable accuracy. As a result, energy consumption and carbon emissions are substantially reduced. Training from scratch consumed approximately 0.58 kWh and emitted 0.22 kg CO$_2$ over 44.5 minutes, while fine-tuning from the pretrained model consumed only 0.46 kWh and emitted 0.18 kg CO$_2$ over 35.6 minutes, a reduction of ~21\% in both energy and emissions. 

We acknowledge that carbon emission values reported by CodeCarbon are based on assumptions about hardware characteristics and regional energy mix, and should therefore be interpreted as approximate estimates. Instead, we use these estimates primarily to enable a consistent relative comparison between training strategies under identical experimental conditions. 

Overall, these results highlight that leveraging pretrained representations provides meaningful environmental and computational benefits, making deep learning for 3D forest analysis more sustainable.

\begin{comment}
\begin{table}[h!]
\centering
\caption{Energy consumption and carbon footprint for different labeling strategies}
\begin{tabular}{lccc}
\hline
\textbf{Metric} & \textbf{0.01\%} & \textbf{1\%} & \textbf{100\%} \\
\hline
Duration (min) & 42.79 & 44.45 & 44.51 \\
Energy consumed (kWh) & 0.5585 & 0.5808 & 0.5813 \\
Emissions (kg CO$_2$) & 0.2128 & 0.2213 & 0.2215 \\
Average GPU Power (W) & 762.20 & 813.66 & 630.25 \\
Average CPU Power (W) & 225.0 & 225.0 & 225.0 \\
Average RAM Power (W) & 70.0 & 70.0 & 70.0 \\
GPU energy share (kWh) & 0.3486 & 0.3628 & 0.3626 \\
CPU energy share (kWh) & 0.1599 & 0.1661 & 0.1667 \\
RAM energy share (kWh) & 0.04998 & 0.05189 & 0.05204 \\
\hline
\end{tabular}
\label{tab:energy_comparison}
\end{table}
\end{comment}

%We compare the computational time of our models against Treeiso, a non-deep learning instance segmentation method. Our model is executed on an Nvidia A100 GPU, achieving significantly faster inference, only XXX seconds to process a single forest region containing approximately XXX  points over XxX m area. In contrast, Treeiso requires XXX seconds to process the same data. It is important to note that Treeiso runs entirely on CPU without requiring specialized hardware. 

%Despite the hardware differences, our results provide a baseline for the computational cost and carbon footprint associated with deep learning-based 3D forest analysis. These findings emphasize the importance of considering both accuracy and sustainability when evaluating machine learning models in ecological applications.

\subsection{Limitations and future directions}
While our approach demonstrates strong performance in data-limited settings, several limitations remain. 

First, our framework does not provide a unified solution across different tasks. Instead, each task is addressed with a task-specific strategy. In the future it will be more important to develop a single self-supervised framework in which a shared encoder can effectively support multiple downstream tasks. Although our self-supervised encoder can be applied across multiple tasks, our experiments demonstrate that additional techniques are still necessary to achieve stronger performance, particularly in few-shot learning scenarios.

For example, instance segmentation in our study still relies on domain adaptation to achieve optimal performance. This highlights a gap in current self-supervised learning methods, which are not yet sufficient to capture instance-level representations independently. Developing self-supervised objectives tailored specifically for instance segmentation remains an important direction, especially to fully leverage large-scale unlabeled data without the need for labeled source domains. 

Second, our models occasionally suffer from false negatives, particularly under complex forest structures. This suggests difficulties in fully separating overlapping or tightly clustered tree instances. To improve this, enhancing offset prediction quality is needed. This can be achieved by introducing an additional branch that learns a richer embedding space for clustering, such as the 5-dimensional embeddings used by \cite{XIANG2024114078}, or by adding extra losses, such as separation or boundary-aware losses.

%%%%%%%%%%%%%%%%%%%%%%%%%%%%%%%%%%%%%
\section{Conclusion}
\label{sec:conclusion}

Accurately extracting structural and species-level information from 3D forest point clouds is essential for precision forestry, biodiversity monitoring, and carbon mapping, yet large-scale manual annotation remains costly and time-consuming. Our results show that few-shot learning strategies, supported by self-supervised pretraining, transfer learning, and domain adaptation, can greatly reduce the need for labeled data while maintaining strong performance.

For instance segmentation, combining self-supervised pretraining with domain adaptation yielded robust results in complex forest structures. In semantic segmentation, self-supervised learning alone improved accuracy and convergence speed, while for tree classification, hierarchical pretraining on broad categories before species-level fine-tuning substantially boosted accuracy. Leveraging pretrained models also potentially reduces training duration, leading to lowering energy consumption and carbon emissions.

Overall, our findings highlight a practical and scalable pathway for 3D forest analysis under real-world constraints. By unifying self-supervised and transfer-based learning, this work moves toward automated, data-efficient forest mapping and bridging the gap between research-grade methods and operational monitoring at scale.

\textbf{Code availability}

The framework, pretrained models, and codes used for training and evaluation are openly available at the following repository: \url{https://github.com/aldinorizaldy/TreeLite3D}

\textbf{Declaration of competing interest}

The authors declare that they have no known competing financial interests or personal relationships that could have appeared to influence the work reported in this paper.

\textbf{Declaration of AI and AI-assisted technologies in the writing process}

During the preparation of this work the authors used OpenAI's ChatGPT v5.0 to improve readability. After using this tool the authors reviewed and edited the content as needed and take full responsibility for the content of the published article.

\textbf{Acknowledgment}

We thank Dr. Benjamin Brede (GFZ Potsdam) for providing the droneborne LiDAR data used in the pretraining stage. This work was also supported by the European Regional Development Fund and the Land of Saxony by providing the high-specification Nvidia A100 GPUs server which we used in our experiments. 

\bibliographystyle{elsarticle-harv}
\bibliography{Reference}

\clearpage

\section*{List of Figure Captions}
\begin{description}
\item[Figure 1] Overview of the framework.
\item[Figure 2] Examples of sampled points and trees under different labeling scenarios. [a,b,c] illustrate uniform label reduction, while [d,e,f] show tree-level label reduction.
\item[Figure 3] Instance segmentation pipeline. The input consists of $N$ points with 3 spatial coordinates ($X,Y,Z$) and $C$ additional attributes (intensity, return number, and number of returns). The backbone encoder extracts 96-dimensional features used by MLP layers to predict point-wise offsets. Points are then translated to shifted coordinates for final instance clustering.
\item[Figure 4] Semantic segmentation pipeline. The input follows the same structure as in instance segmentation pipeline, consisting of $N$ points with 3 spatial coordinates and $C$ additional attributes. The backbone encoder extracts features, which are processed by dual MLP layers to predict point-wise semantic labels.
\item[Figure 5] Tree classification pipeline. Features from the backbone encoder are aggregated to predict final tree class.
\item[Figure 6] Loss monitoring during contrastive pretraining.
\item[Figure 7] Similarity metrics during contrastive pretraining. (a) Positive similarity shows the average cosine similarity between augmented views of the same point cloud sample. (b) Negative similarity shows the average cosine similarity between different point cloud samples.
\item[Figure 8] Visualization of instance segmentation  results on the different forest regions in the FOR-instance dataset.
\item[Figure 9] Performance of instance segmentation under uniform label reduction using stratified sampling on the SCION dataset.
\item[Figure 10] Performance of instance segmentation under tree-level label reduction on the SCION dataset.
\item[Figure 11] Instance segmentation results under different label proportions.
\item[Figure 12] Visualization of semantic segmentation results on the different forest regions in the FOR-instance dataset.
\item[Figure 13] Validation curves of mAP and AP50 under full supervision (a and b) and limited supervision (c and d) during training. Models initialized with self-supervised and domain adaptation pretrained weights (\textcolor{red}{---}) converge faster and achieve higher accuracy, while models trained from scratch (\textcolor{blue}{---}), struggle to optimize in the nearly stages of training and show slower convergence.
\item[Figure 14] Validation curves of loss and mIoU (a and b) during training showing that models initialized with self-supervised pretrained weights (\textcolor{red}{---}) converge faster and achieve better early performance compared to training from scratch (\textcolor{blue}{---}), especially under limited supervision (c and d).
\item[Figure 15] Validation curves of mean accuracy (a) and mean IoU for few-shot learning of tree species classification. Models initialized with pretrained weights (\textcolor{red}{---}) converge faster and achieve highest accuracy during mid-training, while models trained from scratch (\textcolor{blue}{---}), converges much later.

\end{description}

\end{document}